\documentclass[preprint,sort&compress,5p,twocolumn,lefttitle]{elsarticle}

\usepackage{dblfloatfix}

\usepackage{amsmath}
\usepackage{amsfonts}
\usepackage{amssymb}
\usepackage{amsthm}
\makeatletter
\renewenvironment{proof}[1][\proofname]{\par
  \pushQED{\qed}%
  \normalfont \topsep6\p@\@plus6\p@\relax
  \trivlist
  \item\relax
  \ifx#1\empty\else\fi
  \ignorespaces
}{%
  \popQED\endtrivlist\@endpefalse
}
\makeatother

\usepackage{bm}
\usepackage{bbm}
\usepackage{gensymb}
\usepackage{graphicx}
\usepackage{amsmath}
\usepackage{amssymb}
\usepackage{algorithm}
\usepackage[noend]{algpseudocode}
\usepackage{subcaption}
\usepackage{amsfonts}
\usepackage[mode = match]{siunitx}
\usepackage{booktabs}
\usepackage{makecell}
\usepackage{multirow}
\usepackage{upgreek}
\usepackage{pifont}
\usepackage[font=small]{caption}
\usepackage[export]{adjustbox}
\usepackage{tikz}
\usepackage{sidecap} \sidecaptionvpos{figure}{c}
\usepackage{lscape}
\usepackage{threeparttable}

\DeclareMathOperator*{\argmin}{argmin}

\usepackage{hyperref}
\hypersetup{colorlinks,breaklinks,
linkcolor=[rgb]{0.5,0.,0.},
citecolor=[rgb]{0.000,0.427,0.173},
urlcolor=[rgb]{0.031,0.318,0.612}}

\usepackage[nameinlink]{cleveref}
\usepackage{color}
\definecolor{CommentPink}{rgb}{1,0.2,0.5}
\definecolor{CommentBlue}{rgb}{0,0,1}
\definecolor{CommentGreen}{rgb}{0,1,0}

\Crefname{section}{Section}{Section}
\crefname{section}{section}{section}
\Crefname{figure}{Figure}{Figure}
\crefname{figure}{figure}{figure}
\Crefname{equation}{Equation}{Equation}
\Crefname{equation}{equation}{equation}
\crefname{appendix}{}{Appendices}

\crefformat{equation}{(#2#1#3)}

\usepackage[printonlyused,withpage,nolist,nohyperlinks]{acronym}
\begin{acronym}

\acro{2D}{two-dimensional}

\acro{3D}{three-dimensional}

\acro{AHRS}{attitude and heading reference system}

\acro{AUV}{autonomous underwater vehicle}

\acro{COMA}{counterfactual multi-agent policy gradients}

\acro{CPP}{Chinese Postman Problem}

\acro{DoF}{degree of freedom}
\acrodefplural{DoF}[DoFs]{degrees of freedom}

\acro{DVL}{Doppler velocity log}

\acro{FSM}{finite state machine}

\acro{IMU}{inertial measurement unit}

\acro{LBL}{Long Baseline}

\acro{MCM}{mine countermeasures}

\acro{MDP}{Markov decision process}
\acrodefplural{MDP}[MDPs]{Markov decision processes}

\acro{MCTS}{Monte Carlo Tree Search}

\acro{POMDP}{partially observable Markov decision process}
\acrodefplural{POMDP}[POMDPs]{partially observable Markov decision processes}

\acro{PRM}{Probabilistic Roadmap}
\acrodefplural{PRM}[PRM]{Probabilistic Roadmaps}

\acro{RL}{reinforcement learning}

\acro{ROI}{region of interest}
\acrodefplural{ROI}[ROIs]{regions of interest}

\acro{ROS}{Robot Operating System}

\acro{ROV}{remotely operated vehicle}

\acro{SLAM}{simultaneous localization and mapping}

\acro{SSE}{sum of squared errors}

\acro{STOMP}{Stochastic Trajectory Optimization for Motion Planning}

\acro{TRN}{Terrain-Relative Navigation}

\acro{UAV}{unmanned aerial vehicle}
\acrodefplural{UAV}[UAVs]{unmanned aerial vehicles}

\acro{UGV}{unmanned ground vehicle}
\acrodefplural{UAV}[UAVs]{unmanned ground vehicles}

\acro{USBL}{Ultra-Short Baseline}

\acro{IPP}{informative path planning}

\acro{AIPP}{adaptive informative path planning}

\acro{FoV}{field of view}
\acrodefplural{FoV}[FoVs]{fields of view}

\acro{CDF}{cumulative distribution function}

\acro{ML}{maximum likelihood}

\acro{RMSE}{Root Mean Squared Error}
\acro{MLL}{Mean Log Loss}

\acro{GP}{Gaussian Process}
\acrodefplural{GP}[GPs]{Gaussian Processes}

\acro{KF}{Kalman Filter}

\acro{IP}{Interior Point}
\acro{BO}{Bayesian Optimization}

\acro{SE}{squared exponential}

\acro{UI}{uncertain input}

\acro{MCL}{Monte Carlo Localisation}
\acro{AMCL}{Adaptive Monte Carlo Localisation}

\acro{SSIM}{structural similarity index measure}
\acro{PSNR}{peak signal-to-noise ratio}
\acro{LPIPS}{learned perceptual image patch similarity}

\acro{MAE}{Mean Absolute Error}
\acro{RMSE}{root mean squared error}
\acro{AUSE}{Area Under the Sparsification Error curve}
\acro{mIoU}{mean Intersection-over-Union}
\acro{CMA-ES}{covariance matrix adaptation evolution strategy}
\acro{RRT}{rapidly exploring random tree}
\acro{RIG}{rapidly exploring information gathering}

\acro{AL}{active learning}
\acro{MLP}{multi-layer perceptron}
\acro{LSTM}{long short-term memory}
\acro{PPO}{proximal policy optimization}
\acro{CNN}{convolutional neural network}

\acro{IL}{imitation learning}
\acro{GPS}{Global Positioning System}

\end{acronym}

\journal{Robotics and Autonomous Systems}

\begin{document}

\begin{frontmatter}

\title{Approximate Sequential Optimization for Informative Path Planning}

\author[label1]{Joshua Ott\corref{cor1}} \ead{joshuaott@stanford.edu}
\author[label1]{Mykel J. Kochenderfer} \ead{mykel@stanford.edu}
\author[label2]{Stephen Boyd} \ead{boyd@stanford.edu}

\cortext[cor1]{Corresponding author.}

\affiliation[label1]{{Department of Aeronautics & Astronautics, Stanford University}}%
\affiliation[label2]{{Department of Electrical Engineering, Stanford University}}

\begin{abstract}
We consider the problem of finding an informative path through a graph, given initial and terminal nodes and a given maximum path length. We assume that a linear noise corrupted measurement is taken at each node of an underlying unknown vector that we wish to estimate. The informativeness is measured by the reduction in uncertainty in our estimate, evaluated using several Gaussian process-based metrics. We present a convex relaxation for this informative path planning problem, which we can readily solve to obtain a bound on the possible performance. We develop an approximate sequential method where the path is constructed segment by segment through dynamic programming. This involves solving an orienteering problem, with the node reward acting as a surrogate for informativeness, taking the first step, and then repeating the process. The method scales to very large problem instances and achieves performance close to the bound produced by the convex relaxation. We also demonstrate our method's ability to handle adaptive objectives, multimodal sensing, and multi-agent variations of the informative path planning problem.
\end{abstract}


\end{frontmatter}

\section{Introduction}\label{sec:intro}

Consider an agent that is tasked with exploring a large unknown environment. The agent must collect measurements to build an accurate representation of the state of the environment. The agent has a finite set of resources and is therefore constrained by time, battery life, or fuel capacity. As a result, it must plan a path to maximize the amount of information acquired from its noisy observations while remaining within the allowed resource budget. Given a maximum path length $L^\text{max}$, the objective is to compute a path of length at most $L^\text{max}$ that maximizes the gain in information over the prediction variables. This is known as the informative path planning (IPP) problem and involves choosing a subset of sensing locations from the environment. The IPP problem can be seen as an extension of the sensor selection problem where future sensing locations are constrained by previous sensing locations. These additional path feasibility constraints increase the complexity of the problem, making it more challenging than the original sensor selection problem \cite{joshi2008sensor, dutta2022informative}.    

A variety of real-world problems can be formulated as IPP problems, such as planetary rover exploration, search and rescue, and environmental monitoring. IPP merges the fields of robotics, artificial intelligence, and spatial data analysis. It carries substantial implications for a myriad of applications, where efficient data collection from a potentially hazardous or inaccessible environment is essential. However, the problem presents significant challenges. The IPP problem is known to be NP-hard due to its combinatorial nature with a vast search space for potential solutions \cite{meliou2007nonmyopic}. Successful solutions typically require reasoning about the entire trajectory; myopic approaches often result in inefficient paths, causing the agent to repeat exploration of the same areas without using the full resource budget \cite{morere2017sequential, popovic2020informative, vashisth2024deep, yu2022efficient, jakkala2023multi}.

The IPP problem is closely related to the optimal design of experiments and sensor selection. \citeauthor{joshi2008sensor} presented a convex relaxation of the sensor selection problem that involves choosing a set of $k$ sensor measurements, from a set of $m$ possible or potential sensor measurements, that minimizes the error in estimating some parameters \cite{joshi2008sensor}. The sensor selection problem is closely related to the IPP problem. However, sensor selection does not consider the movement constraints of the agent so the rounding approach presented by \citeauthor{joshi2008sensor} is not directly applicable to the IPP problem. \citeauthor{krause2008near} also considered the sensor placement problem and presented several approximate solution methods \cite{krause2008near}. \citeauthor{singh2009efficient} introduced an approximate algorithm for both the single and multi-agent IPP problems using Gaussian processes as the underlying environment model, but they do not consider the adaptive IPP problem \cite{singh2009efficient}. \citeauthor{strawser2022motion} used hybrid search to solve risk-bound goal-directed planning problems, focusing on complex 3D environments and the modeling of trajectory risk \cite{strawser2022motion}. \citeauthor{backstrom2021cost} focused on approximability for cost-optimal planning \cite{backstrom2021cost}. 

\citeauthor{dutta2022informative} introduced a mixed-integer formulation for the discrete IPP problem in random fields and aimed to optimize over the collected measurement subset and all linear estimators \cite{dutta2022informative}. The approach presented by \citeauthor{dutta2022informative} can produce globally optimal solutions on relatively small graphs; however, the mixed integer program fails to scale to larger graph sizes which is often necessary in practice for planning at higher resolutions \cite{dutta2022informative}. \citeauthor{bostrom2018global} formulated the problem as a mixed integer semidefinite program and used a receding horizon optimal control approach based on the information filter to produce a solution \cite{bostrom2018global}. \citeauthor{song2015trajectory} used a receding horizon control strategy with a discrete set of motion primitives to evaluate the objective and then select the best candidate, execute the action, update the Gaussian process belief, and then replan \cite{song2015trajectory}.

\textcolor{black}{In addition to the discrete path planning methods discussed, continuous informative path planning approaches have also been explored. These methods focus on optimizing continuous paths rather than discrete sensing locations, which is crucial for certain real-world applications where smooth trajectories and compliance with motion constraints are required. \citeauthor{hitz2017adaptive} used an evolutionary strategy to optimize a parameterized path in continuous space \cite{hitz2017adaptive}. \citeauthor{francis2019occupancy} presented a constrained Bayesian optimization approach to plan continuous paths that inherently satisfy motion and safety constraints while balancing the reward and risk associated with each path \cite{francis2019occupancy}. \citeauthor{asgharivaskasi2022active} proposed a differentiable approximation of the Shannon mutual information between a grid map and ray-based observations, enabling gradient ascent optimization in the continuous space of SE(3) sensor poses \cite{asgharivaskasi2022active}.}

Another common approach is to formulate the IPP problem as a partially observable Markov decision process (POMDP). \citeauthor{marchant2014sequential} formulated the IPP problem as a POMDP and solved it using sequential Bayesian optimization through Monte Carlo tree search with upper confidence bound for trees (MCTS-UCT) \cite{marchant2014sequential}. This work has been extended to modify the reward function for achieving monitoring behavior that exploits areas with high gradients, and for additional reasoning over continuous action spaces through Bayesian optimization \cite{morere2017sequential, morere2018continuous}. \citeauthor{fernandez2022informative}, on the other hand, used partially observable Monte Carlo planning (POMCP) with Gaussian process beliefs for estimating quantiles of the underlying world state \cite{fernandez2022informative}. Their methodology proposes sample locations for a team of scientists to investigate following their exploration completion. \citeauthor{ruckin2022adaptive} presented a hybrid approach that combines tree search with an offline-learned neural network to predict informative sensing actions \cite{ruckin2022adaptive}. \citeauthor{ott2023sequential} focused on the multimodal sensing extension of the IPP problem where the agent has multiple sensors to choose from and solved it with tree search \cite{ott2023sequential}. 

\textcolor{black}{These approaches all produce reasonable solutions on graphs with a relatively small number of nodes; however, they struggle to scale to large graphs with higher resolution which is necessary for many real-world applications. Additionally, MIPs are not well-suited to handle adaptive objectives where the objective function depends on the history of measurement values received. Our results demonstrate that our approach is able to scale to larger problem instances for the IPP problem, while also handling adaptive objectives such as the expected improvement metric, and accounting for multimodal sensing and multi-agent variations of the IPP problem \cite{choudhury2020adaptive, ott2023sequential}. We directly compare our approach with those of \citeauthor{dutta2022informative} and \citeauthor{ott2023sequential} \cite{dutta2022informative, ott2023sequential}.}

\textcolor{black}{The key contributions of this work are the following:}
\begin{enumerate}
    \item \textcolor{black}{We present a convex relaxation of the informative path planning problem.}

    \item \textcolor{black}{We introduce an approximate sequential optimization solution that outperforms existing IPP methods.}

    \item \textcolor{black}{We provide a bound on the optimal solution to the IPP problem and show that while our method produces a suboptimal solution, often the optimality gap is small.}

    \item \textcolor{black}{We extend our method to handle adaptive objectives such as the expected improvement metric, as well as other variations of the IPP problem including multimodal sensing and multi-agent scenarios \cite{kochenderfer2019algorithms}.}

    \item \textcolor{black}{We release our implementation as an open-source software package for use and further development by the community.}\footnote{https://github.com/sisl/InformativePathPlanning}
\end{enumerate}

The remainder of this paper is organized as follows: \cref{sec:ipp} presents the IPP problem, \cref{sec:mi_convex} formulates the IPP problem with an exact mixed integer convex formulation and presents a convex relaxation,  \cref{sec:cocp} introduces our approximate sequential path optimization solution, \cref{sec:extensions_and_variations} extends our method to consider additional adaptive objectives and multimodal sensing variations of the original problem, \cref{sec:results} presents our empirical results along with a bound on the optimality of our approximate solution, and \cref{sec:conclusion} concludes our work.

\section{Informative Path Planning}\label{sec:ipp}
\subsection{Feasible Paths}
We model the environment as a directed graph with vertices $V = \{ 1, \dots, n\}$ and edges $\mathcal{E} \subseteq V \times V$. %
Our goal is to find a feasible path $p$ from the start vertex $s$ to the goal vertex $g$ where we assume that such a feasible path exists. We denote a path of length $L$ as $p = ( i_1, i_2, \dots, i_{L} )$ where $i_1 = s$ and $i_L = g$. A feasible path is one that starts at $s$ and ends at $g$ while remaining within the given maximum path length $L^\text{max}$; that is, $L \leq L^{\text{max}}$. %

In practice, the vertices of the graph correspond to the locations in the environment we are interested in exploring. Additionally, environmental obstacles can be incorporated by removing edges in the graph. However, these practical considerations do not alter the generality of our formulation.

\subsection{Sensing Along a Path}
We're interested in estimating a vector $x \in \mathbf{R}^m$ from a set of linear measurements, corrupted by additive noise, 
\begin{equation}
    y_i = a_i^Tx + \nu_i
\end{equation} where $x \in \mathbf{R}^m$ is a vector of parameters to estimate, each $\nu_i \sim \mathcal{N}(0, \sigma^2)$ is independent and identically distributed additive noise, and $a_i \in \mathbf{R}^{m}$ characterizes the measurements. 

We assume we have some prior over $x$ with known mean and covariance $\mathcal{N}(\bar{x}, \Sigma_x)$. Here, $x$ can represent subsurface temperature, ice concentration, algae density, or a variety of other spatially varying phenomena.

We assume that at each node $i_k$ along a path $p$ we make measurement $y_k$. The measurements received along a path $p$ of length $L$ are then given by \begin{equation}
    y_k = a_{i_k}^Tx + \nu_{i_k} \quad k = 1, \dots, L.
\end{equation}

The maximum a posteriori estimate of $x$ is then \begin{equation}
\hat{x} = \bar{x} + \left( \sigma^{-2} \sum_{k=1}^L a_{i_k} a_{i_k}^T + \Sigma_x^{-1} \right)^{-1} \sigma^{-2} \sum_{k=1}^L (y_k - \bar{y}_k)a_{i_k}  \label{eq:gp_mean} 
\end{equation} where $\hat{x} \in \mathbf{R}^m$ with covariance \begin{equation}
\Sigma = \left( \sigma^{-2}  \sum_{k=1}^L a_{i_k} a_{i_k}^T + \Sigma_x^{-1} \right)^{-1}
\label{eq:gp_var}
\end{equation} where $\bar{x}$ is the prior mean and $\bar{y}_k = a_{i_k}^T \bar{x}$. Note that this is exactly the Gaussian process mean and covariance as shown in \cref{sec:gp_equivalence}.

\subsection{Measures of Informativeness} \label{ss:measure_informativeness}
Three common scalar measures of the informativeness of a path are: \begin{enumerate}
    \item $\mathbf{tr}(\Sigma) = \mathbf{E}\left( \lVert  x-\hat{x} \rVert^2 \right)$, which corresponds to the mean squared error in estimating the parameter $x$.
    \item $-\mathbf{tr}(\Sigma^{-1})$, which is the negative trace of the information matrix. We include the negative here since we are interested in minimizing each of these objectives in order to maximize the informativeness of the path.
    \item $\operatorname{logdet}(\Sigma)$, which is proportional to the log volume (or mean radius) of the resulting confidence ellipsoid.
\end{enumerate} Note that each scalar measure of informativeness is a function of the eigenvalues of $\Sigma$. We will denote the measure of informativeness of a path as $\phi(p)$ where smaller values of $\phi(p)$ correspond to more informative paths. \textcolor{black}{Other measures of informativeness such as adaptive objectives will be discussed in \cref{ss:additional_objectives} and we also provide a formulation of the mutual information in \cref{sec:mutual_information}.}

\subsection{Informative Path Planning Problem}
The informative path planning (IPP) problem seeks to find the maximally informative feasible path $p$ from the set of all feasible paths $\mathcal{P}$. The IPP problem can be stated as follows \begin{align} 
\text{minimize} \quad & \phi(p) \nonumber \\
\text{subject to} \quad & p \in \mathcal{P} \label{eq:ipp_problem} \end{align} where the variables are the nodes in the path $p$ and the data consists of the graph, the measurement characterization $a_i$, the measurement variance $\sigma^2$, and the prior covariance $\Sigma_x$. This is a combinatorial problem since the variable is a path.

\subsection{Relation to A and D-Optimal Experiment Design}
Depending on the information objective used, the informative path planning problem is closely related to the experiment design problem. The experiment design problem involves choosing $c$ out of $h$ available sensors. For each of the selected $c$ sensors, we also need to determine how many times it should be used while respecting the constraints on the number of total allowable uses \cite{joshi2008sensor}. A-optimal experiment design seeks to minimize $\mathbf{tr}(\Sigma)$ as the information objective while D-optimal design seeks to minimize $\operatorname{logdet}(\Sigma)$. While not commonly named, we introduce B-optimal design which seeks to minimize $-\mathbf{tr}(\Sigma^{-1})$ and we show in \cref{ss:approx_a_opt_objective} that it proves to be a powerful surrogate for both A and D-optimal objectives. 

Informative path planning focuses on selecting specific sensing locations from the entire environment. While it builds upon the sensor selection problem, it introduces additional constraints: future sensing locations can only be chosen if they are adjacent to previously selected locations. These path feasibility constraints introduce additional combinatorial complexity, making the problem more challenging than the original sensor selection task.

We will refer to the $\mathbf{tr}(\Sigma)$, $\operatorname{logdet}(\Sigma)$, $-\mathbf{tr}(\Sigma^{-1})$ objectives as the A-IPP, D-IPP, and B-IPP objectives respectively. For the A-IPP objective we seek to minimize $\phi(p) = \mathbf{tr}(\Sigma)$, for the B-IPP objective we seek to minimize $\phi(p) = -\mathbf{tr}(\Sigma^{-1})$, and for the D-IPP objective we seek to minimize $\phi(p) = \operatorname{logdet}(\Sigma)$.

\section{Mixed Integer Convex Formulation}\label{sec:mi_convex}
We now introduce a Boolean matrix $z \in \mathbf{R}^{n \times n}$ which we will use to parameterize a path $p$. Here, $z_{ij} = 1$ if the edge from node $i$ to node $j$ is in the path $p$. We must have, \begin{equation}
    z_{ij} = 0 \quad \mathrm{for} \quad (i,j) \notin \mathcal{E}.
\end{equation} That is, the sparsity pattern of $z$ is the same as that of the environment graph. 

Using the Boolean parametrization of $z$ we can define a feasible path. A feasible path must remain within the given maximum path length $L^\text{max}$: %
\begin{equation}
    \sum_{i=1}^{n-1} \sum_{j \in N_{out}(i)} z_{ij} \leq L^\text{max}
    \label{eq:z_budget}
\end{equation} where $N_{out}(i)$ denotes the graph neighbors leaving node $i$ and $N_{in}(i)$ denotes the graph neighbors entering node $i$.  

If we take node $1$ and $n$ as the start and goal nodes respectively, then we know the start and end nodes must exist on the path meaning that one edge going out of the start node and one edge going into the goal node must be in the path: \begin{equation}
    \sum_{i \in N_{out}(1)} z_{1i} = \sum_{j \in N_{in}(n)} z_{jn} = 1. \label{eq:z_start_goal}
\end{equation} Similarly, to enforce the start and termination of the path, we do not allow any edges to enter the starting node or leave the goal node: \begin{equation}
    \sum_{i \in N_{in}(1)} z_{i1} = \sum_{j \in N_{out}(n)} z_{nj} = 0.
    \label{eq:z_termination}
\end{equation}

To ensure the connectivity of the path and that every node is visited at most once, we enforce \begin{equation}
    \sum_{\substack{j \in N_{out}(i) \\ j\neq 1}} z_{ij} = \sum_{\substack{k \in N_{in}(i) \\ k \neq n}} z_{ki} \leq 1 \quad  i = 2,\ldots,n-1.
    \label{eq:z_connect}
\end{equation}

To prevent any subtours from occurring, we use the Miller-Tucker-Zemlin constraints for subtour elimination \cite{miller1960integer}. To do so, we introduce the variable $u$ which enforces that node $j$ must follow node $i$ if edge $(i,j)$ is in the path $p$:\begin{align}
    u_i - u_j + 1 \leq (n-1)(1-z_{ij}) \quad & 2 \leq i \neq j \leq n \label{eq:z_sub1}\\
    2 \leq u_i \leq n \quad & 2 \leq i \leq n \label{eq:z_sub2}\\
    u_1 = 1. \label{eq:z_sub3}
\end{align}
Finally, to enforce the integrality of the Boolean matrix $z$ we have: \begin{equation}
    z_{ij} \in\{0,1\} \quad i = 1,...,n. \label{eq:z_int}
\end{equation}

The constraints provided in \Cref{eq:z_budget} - \Cref{eq:z_int} fully define the space of feasible path parameterizations $\mathcal{Z}$. We then have \begin{equation}
\Sigma_z = \left( \sigma^{-2} \sum_{i=1}^n \left( \sum_{j \in N_{out}(i)} z_{ij} \right) a_i a_i^T + \Sigma_x^{-1} \right)^{-1}
\label{eq:sigma_z}
\end{equation} and can write our parameterized objective as $\phi(z)$. We then have the following convex mixed integer problem: \begin{align}
\text{minimize} \quad & \phi(z)  \nonumber \\
\text{subject to} \quad & z \in \mathcal{Z}. \label{eq:exact_ipp} \end{align} While it is not immediately obvious, the $\phi(z)$ objective is convex in both the A, B, and D-IPP objectives introduced in \cref{ss:measure_informativeness} as discussed in section 4.6.2 of \citeauthor{boyd2004convex}. The problem would be entirely convex if it were not for the integer constraints on $z$ in \Cref{eq:z_int}.

\subsection{Convex Relaxation} \label{ss:convex_relaxation}
The exact mixed integer convex problem in \Cref{eq:exact_ipp} can be converted to a fully convex problem by relaxing the integer constraints on $z$. Specifically, we replace the $z_{ij} \in\{0,1\}$ constraint with $0 \leq z_{ij} \leq 1$. 

The solution to the relaxed problem is denoted as $\Tilde{z}$ and provides a lower bound $\ell$ on the optimal solution $z^{\star}$. The reason for this is that the feasible set for the relaxed problem contains the feasible set for the original problem and therefore the optimal value of the original problem cannot be smaller than that of the relaxed problem \cite{boyd2004convex}. In the case where $\Tilde{z}$ is Boolean, then we can say that $z^{\star} = \Tilde{z}$.

\section{Approximate Sequential Path Optimization}\label{sec:cocp}
The exact formulation provided in \cref{sec:mi_convex} is a mixed-integer convex problem (MICP). For relatively small problems, these MICPs are solvable, and we provide an implementation using the Pajarito solver in Julia.\footnote{https://github.com/jump-dev/Pajarito.jl} However, the MICPs become prohibitively expensive computationally as $n$ grows. The relaxed version of the problem is able to scale to much larger problems $(n > 10000)$.

\begin{figure*}[t]
\centering
    {\includegraphics[width=0.85\textwidth]{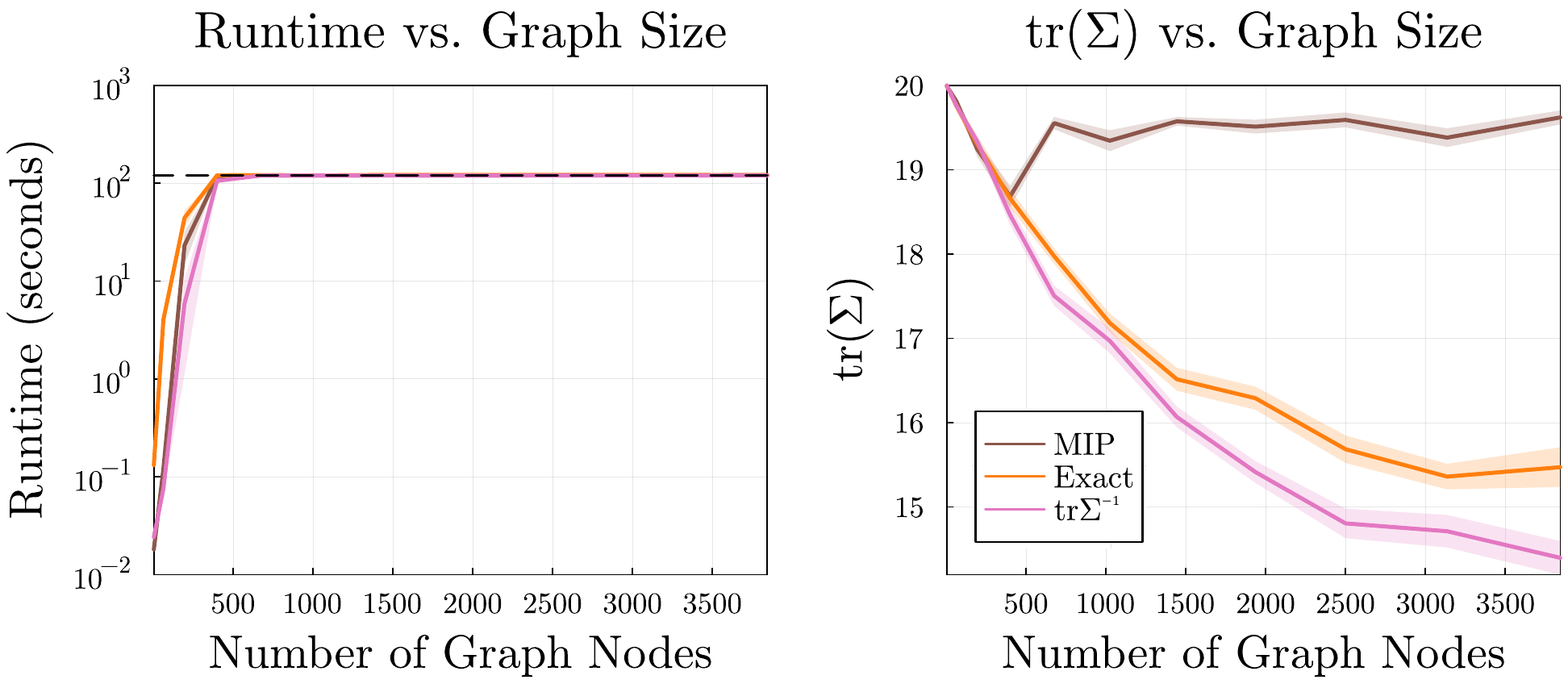}}    
  \caption{Comparative analysis of computational runtime (left) and the objective function $\textbf{tr}(\Sigma)$ (right) as functions of the graph size. Three methods are compared: the mixed integer program formulation \cite{dutta2022informative}, the exact formulation from \cref{eq:exact_ipp}, and the exact method using the B-IPP $\textbf{tr}(\Sigma^{-1})$ objective. Each curve shows the average runtime and objective value respectively with the standard error reported over 25 different simulations. The dashed line on the left indicates the 120 second runtime constraint imposed on all methods.} \label{fig:mip_exact_tr_comparison}
\end{figure*}

We now introduce our method, which we refer to as approximate sequential path optimization (ASPO), which achieves similar performance to that of the exact mixed integer solution while maintaining a similar speed and scalability of the relaxed version. 

Our approximate method begins with the agent at the starting vertex assuming no measurements have been taken yet. For each of the $n$ vertices in the environment, we compute $r_j$, which is the estimated value of visiting node $j$ conditioned on the previously visited nodes. For example, \begin{equation}
    r_j = -\phi((p_t, j))
\end{equation} %
which is the objective value if node $j$ was added to the current executed path $p_t$ at time $t$. Note the negative sign since we seek to minimize the $\phi(p)$ objective which corresponds to maximizing the informativeness of a path.

The objective is to find a path from $s$ to $g$ that maximizes the total reward collected, subject to a constraint on the total path length not exceeding $L^\text{max}$. This is known as the Orienteering Problem (OP) \cite{vansteenwegen2011orienteering, gunawan2016orienteering} and is formally stated as: \begin{align}
\text{maximize} \quad & \sum_{i=1}^{n-1} \sum_{j \in N_{out}(i)} z_{ij} r_{j} \nonumber \\
\text{subject to} \quad & z \in \mathcal{Z}. \label{eq:approximate_convex_policy} \end{align} 

Solving the exact OP is NP-hard \cite{gunawan2016orienteering}; however, we can use dynamic programming to solve an approximate version of the OP efficiently. Let $U_{i, b}$ denote the value for each node $i$ and remaining path length $b$. Then we have \begin{equation}
    U_{i, b} = r_i + \max_{j \in N_{out}(i)} (U_{j, b - 1}). \label{eq:dynamic_programming}
\end{equation} The path $\hat{p}_t$ is then constructed by starting at $s$ and following the maximum $U_{j,b}$ value for $j \in N_{out}(i)$. The first step, or first $h$ steps, of the candidate $\hat{p}_t$ are executed and the problem is then re-solved with the newly executed path and the updated values for $r_j$.  

Sequentially constructing the path in this receding horizon approach allows for efficient solutions to relatively large problems. The key reason for this is that by approximating the value of node $j$ we are better able to isolate valuable regions of the environment to visit. While this assumption assumes that the agent can essentially teleport from its current location to anywhere in the reachable environment, it proves to be useful from the computational savings it provides and allows for many successive iterations of the problem to construct the path in a receding horizon fashion. As a result, the approximation of the value $r_j$ improves as more of the path is executed. This heuristic proves to be valuable because it allows the problem to remain tractable for large $n$.

\subsection{Optimality Gap \label{ss:optimality_gap}}
The path $p$ obtained from our approximate sequential path optimization can be used to construct the matrix $\hat{z}$ which gives an upper bound $u$ on the value of the IPP problem. From \cref{sec:mi_convex}, we know that the relaxed problem provides a lower bound $\ell$ on the optimal solution. Therefore, the difference between the upper and lower bounds gives us the optimality gap \begin{equation}
    \delta = \frac{1}{m} \left( u - \ell \right) = \frac{1}{m} \left( \phi(\hat{z})  - \phi(\Tilde{z}) \right).
    \label{eq:delta}
\end{equation} 

For D-IPP we can express the gap as the ratio of mean radii $e^{\delta}$. We know that $\delta \geq 0$. If $\delta = 0$ then $\hat{z}$ is the optimal solution for the IPP problem $\hat{z} = z^{\star}$. We can say that our approximate sequential path optimization solution is no more than $\delta$-suboptimal. We present empirical results with evaluations of the optimality gap in \cref{ss:results_optimality_gap}.

\section{Extensions and Variations}\label{sec:extensions_and_variations}
\begin{figure*}[t]
\centering
    {\includegraphics[width=0.85\textwidth]{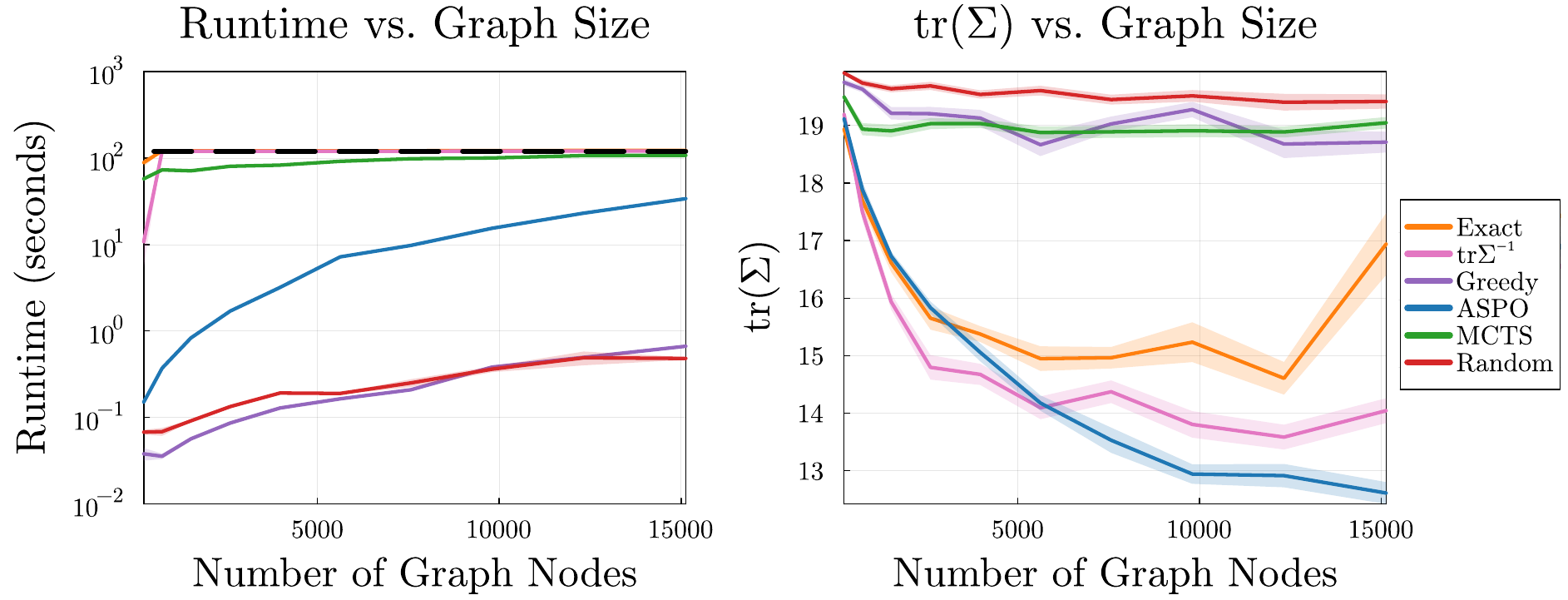}} 
  \caption{Runtime and A-IPP objective as a function of graph size for the six methods considered. Each curve shows the average runtime and objective value respectively with the standard error reported over 25 different simulations. The dashed line on the left indicates the $120$ second runtime constraint imposed on all methods.} \label{fig:a-optimal}
\end{figure*}

The IPP problem has several extensions inspired by numerous real-world applications. We introduce three such extensions below that have been widely studied in the literature. We demonstrate the flexibility of our approximate sequential path optimization approach in handling these extensions which are not easily handled by mixed integer formulations.  

\subsection{Edge Weights}\label{ss:edge_weights}
Our formulation provided thus far has used a constraint on the maximum path length. The IPP problem can also be formulated where each edge \((i, j) \in \mathcal{E}\) is assigned a weight \( d_{ij} \), where \( d_{ij} \) denotes the weight of the edge from vertex \(i\) to vertex \(j\). The maximum path length constraint then becomes a budget constraint where the path cost must not exceed a given budget $B$. That is, for each node $i_k$ in path $p$ we have $(i_k, i_{k+1}) \in \mathcal{E}$ and $\sum_{k=1}^{L-1} d_{i_k, i_{k+1}} \leq B$. The constraint in \cref{eq:z_budget} is then given by \begin{equation}
    \sum_{i=1}^{n-1} \sum_{j \in N_{out}(i)} z_{ij} d_{ij} \leq B. 
\end{equation} 

\subsection{Additional Objectives}\label{ss:additional_objectives}
The approximate sequential path optimization formulation we presented above allows for the inclusion of adaptive objectives, like lower confidence bound and expected improvement, that depend on the received sample values \cite{kochenderfer2019algorithms}. Incorporating adaptive objectives is typically referred to as the adaptive informative path planning (AIPP) problem \cite{choudhury2020adaptive, ott2023sequential}. These adaptive objectives cannot be handled by mixed integer formulations because the objectives require reacting to new observations as they become available. In other words, the belief about the true state of the environment changes in response to the values of the observations. So while A and D-optimal experiment design can be computed a priori, adaptive objectives must be reasoned about in a receding horizon fashion.

 The expected improvement metric at node $i$ is given by: \begin{equation}
     \mathbb{E}[I_{p}]_j = (y_{\text{min}} - \hat{x}_{j}) \Phi \left( \frac{y_{\text{min}} - \hat{x}_{j}}{\hat{\sigma}_j}\right) + \hat{\sigma}_j^2 \mathcal{N}(y_{\text{min}} \mid \hat{x}_j, \hat{\sigma}_j^2) \label{eq:expected_improvement}
 \end{equation} 
 where $y_{\text{min}}$ is the current minimum value found so far, and $\Phi (\cdot)$ is the cumulative distribution function \cite{kochenderfer2019algorithms}. %
 Our problem then becomes \begin{align}
\text{minimize} \quad & \sum_{j=1}^m \mathbb{E}[I_{p}]_j \nonumber \\
\text{subject to} \quad & z \in \mathcal{Z} \end{align} %
where the constraints are equivalent to those in \cref{eq:exact_ipp}. Intuitively, we want to minimize the net amount of expected improvement which corresponds to being very certain that we have found the minimum value in the environment. 

This is a non-convex objective and is not easily handled by current optimizers. However, our approximate sequential path optimization approach can handle this objective by computing $r_j = \mathbb{E}[I_{p}]_j$ for each of the $m$ prediction vertices in the environment with the current values of $\hat{x}$ and $\Sigma$ obtained from the current path $p_t$. Note that we want to visit locations in the environment with high expected improvement to minimize the net expected improvement left in the environment so there is no negative sign in the expression for $r_j$

\begin{figure*}[t]
\centering
    {\includegraphics[width=0.85\textwidth]{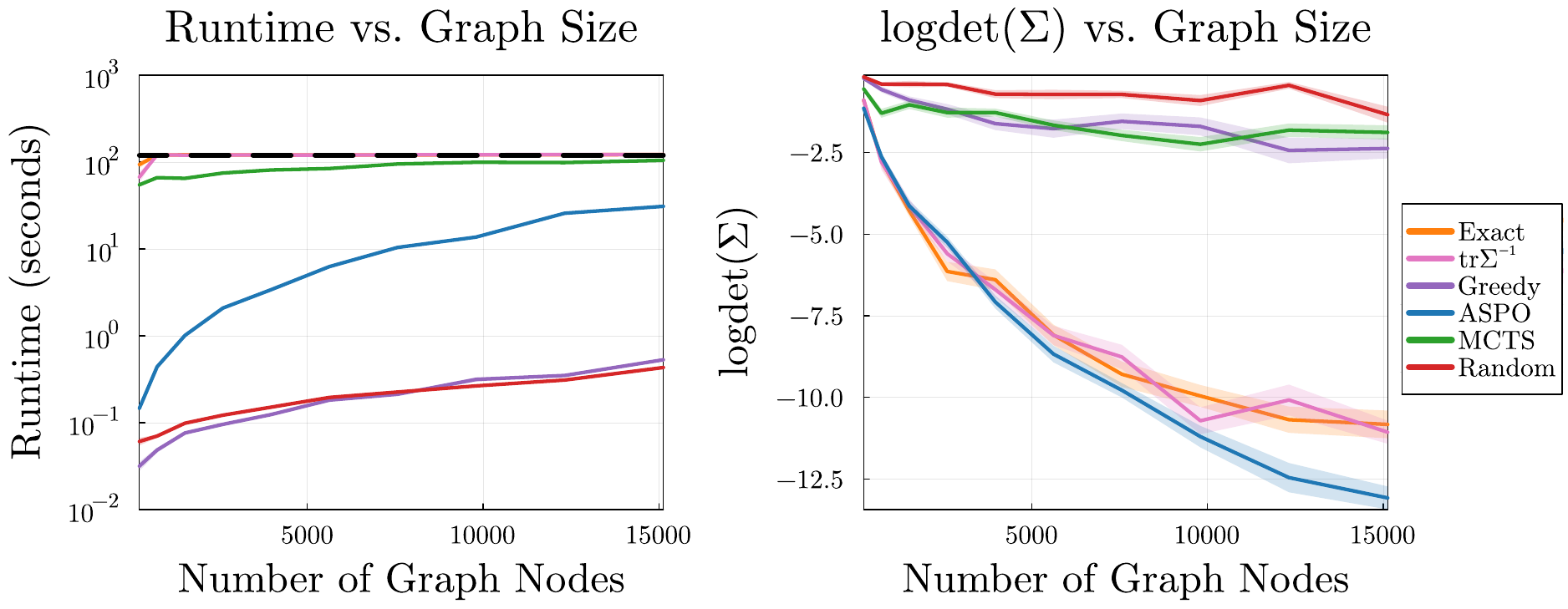}} 
  \caption{Runtime and D-IPP objective as a function of graph size for the six methods considered. Each curve shows the average runtime and objective value respectively with the standard error reported over 25 different simulations. The dashed line on the left indicates the $120$ second runtime constraint imposed on all methods.} \label{fig:d-optimal}
\end{figure*}

\subsection{Multimodal Sensing}\label{ss:multimodal}
Up to this point, we have considered that each measurement taken by the agent is characterized by the same noise distribution $\mathcal{N}(0, \sigma^2)$. We can also consider a further extension of the problem to multimodal sensing domains where each measurement location also involves choosing a sensor type. The sensor type directly influences the variance $\sigma_i^2$ at the measured location. Adaptive informative path planning with multimodal sensing (AIPPMS) considers the case where the agent can choose between multiple sensing modalities (i.e. multiple sensors are available to choose from) \cite{choudhury2020adaptive, ott2023sequential}. This extension adds additional complexity by requiring that the agent reasons about the cost-benefit trade-off associated with the different sensing modalities.

To solve the AIPPMS problem, we first solve the AIPP problem with our approximate sequential path optimization approach and the respective objective presented in \cref{sec:cocp}. We then use the resulting path as the starting point for the multimodal sensor selection problem. The multimodal sensor selection problem is a variant of the sensor selection problem presented by \citeauthor{joshi2008sensor}. In this case, we assume we have $k$ high-quality sensors to choose from with corresponding $\sigma_i$ values ranging from $\sigma_{\text{min}}$ to $\sigma_{\text{max}}$. We constrain the multimodal sensor selection problem to only consider sensor locations along the path $p$. The multimodal sensor selection problem can then be stated as: \begin{align}
\text{minimize} \quad & \phi(s)  \nonumber \\
\text{subject to} \quad & \textbf{1}^Ts = k \nonumber \\
& 0 \leq s_i \leq 1 \nonumber \\
& s_i = 0 \quad \forall i \notin p \label{eq:sensor_selection} \end{align} 
where $s_i$ is the relaxed decision variable corresponding to the importance of placing high-value sensors at location $i$ with \begin{equation}
    \Sigma_s^{-1} = \left(\sigma_{\text{min}}^{-2} \sum_{i=1}^n s_i a_i a_i^T + \Sigma_p^{-1} \right)
\end{equation} and \begin{equation}
    \Sigma_p^{-1} = \left( \sigma_{\text{max}}^{-2} \sum_{i=1}^n \left( \sum_{j \in N_{out}(i)} \hat{z}_{ij} \right) a_i a_i^T + \Sigma_x^{-1} \right)
\end{equation} and $\phi(s)$ is the corresponding IPP objective from \cref{ss:measure_informativeness} using the decision variable $s$. Note that $\Sigma_p$ is the posterior covariance matrix from our path which has now become our prior for maximum a posteriori sensor selection. Additionally, notice that the individual values of $\sigma_i$ are not used. We only require the bounds on their values. Similar to the procedure presented by Joshi and Boyd, once we solve this relaxed problem we then take the largest $k$ values of $s_i$ and assign them sensor values starting at $\sigma_{\text{min}}$ corresponding to the largest $s_i$ and proceeding onwards to $\sigma_{\text{max}}$. In this way, we can think of each $s_i$ as corresponding to the importance of that particular location. Locations with greater importance should be allocated greater resources. In the case where there are only two sensors to choose from and one needs to assign $h$ measurement locations to the more accurate sensor $\sigma_{\text{min}}$, we can simply set the largest $h$ values of $s_i$ to $\sigma_{\text{min}}$. There are other versions of the multimodal sensor selection problem that can be formulated in a similar fashion.

\subsection{Multi-agent IPP}
Consider a set of $M$ agents \( \mathcal{A} = \{A_1, A_2, \dots, A_M\} \). Each agent \( A_i \) is tasked with traversing the graph to collect measurements. The agents share their measurements, forming a collective measurement vector. For agent \( A_j \), its path is \( p^j = \{ i_1^j, i_2^j, \dots, i_{L_j}^j \} \), where \( L_j \) is the length of agent \( A_j \)'s path. Along each path \( p^j \), agent \( A_j \) makes measurements. The measurement at node \( i_k^j \) by agent \( A_j \) is represented as \begin{equation}
    y_k^j = a_{i_k^j}^Tx + \nu_{i_k^j} 
\end{equation} which we can use to construct a unified measurement vector \( Y \) that combines measurements from all agents \begin{equation}
    Y = \{ y_1^1, \dots, y_{L_1}^1, y_1^2, \dots, y_{L_2}^2, \dots, y_1^M, \dots, y_{L_M}^M \} .
\end{equation}

The estimation of \( x \) is now based on the collective measurements from all agents similar to the sequential allocation method proposed by \citeauthor{singh2009efficient} for multi-agent exploration \cite{singh2009efficient}. The maximum a posteriori estimate from all agents measurements is: \begin{equation}
    \hat{x} = \bar{x} + \left( \sigma^{-2} \sum_{j=1}^{M} \sum_{k=1}^{L_j} a_{i_k^j} a_{i_k^j}^T + \Sigma_x^{-1} \right)^{-1} \sigma^{-2} \sum_{j=1}^{M} \sum_{k=1}^{L_j} (y_k^j - \bar{y}_k^j)a_{i_k^j} 
\end{equation} with covariance \begin{equation}
    \Sigma = \left( \sigma^{-2} \sum_{j=1}^{M} \sum_{k=1}^{L_j} a_{i_k^j} a_{i_k^j}^T + \Sigma_x^{-1} \right)^{-1} 
\end{equation} where \( \bar{y}_k^j = a_{i_k^j}^T \bar{x} \).

Similarly, we now have for each agent \( A_j \), a Boolean matrix \( z^j \in \mathbf{R}^{n \times n} \) to parameterize its path \( p^j \). Each \( z^j_{ik} = 1 \) if the edge from node \( i \) to node \( k \) is part of agent \( A_j \)'s path, and 0 otherwise. \( \Sigma \) now takes into account the paths of all agents: \begin{equation}
    \Sigma = \left( \sigma^{-2} \sum_{j=1}^{M} \sum_{i=1}^n \left( \sum_{k \in N_{out}(i)} z^j_{ik} \right) a_i a_i^T + \Sigma_x^{-1} \right)^{-1}.
\end{equation}

\begin{figure*}[t]
\centering
    {\includegraphics[width=1\textwidth]{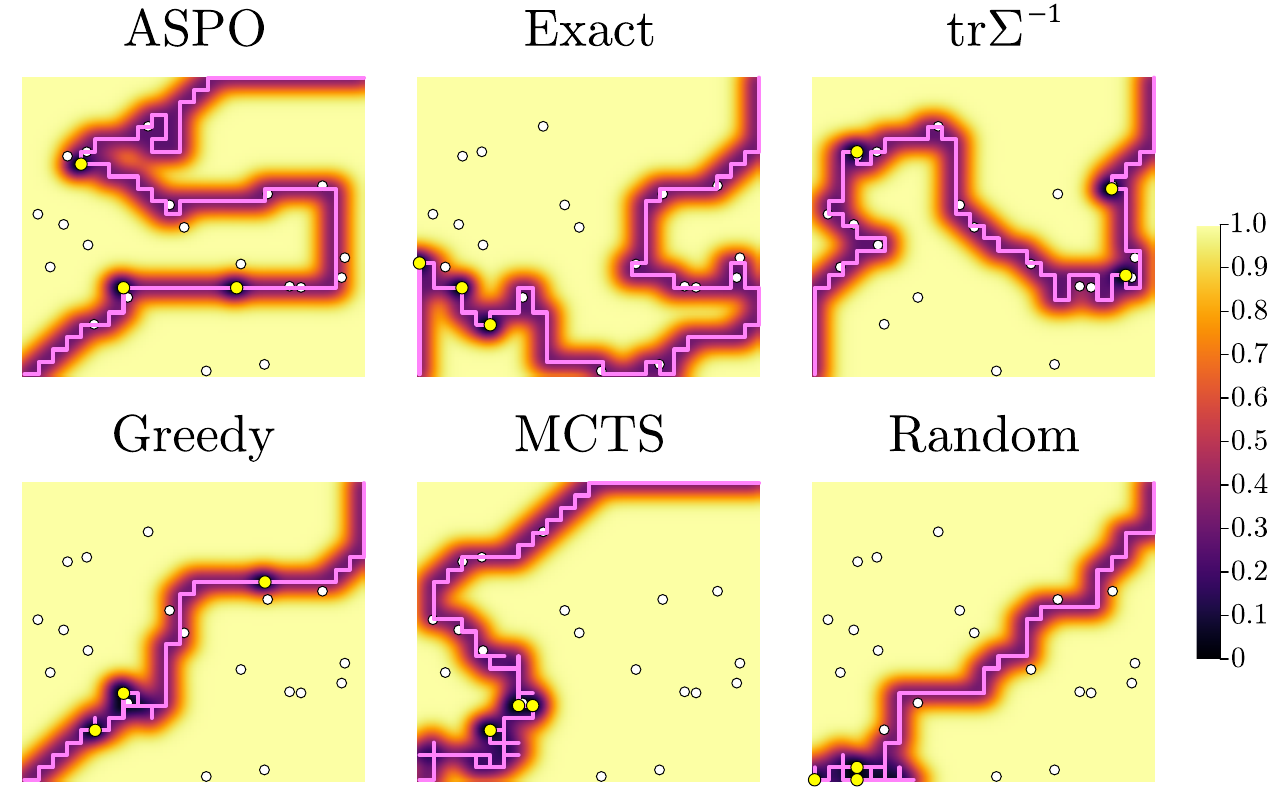}}    
  \caption{Examples of trajectories produced by each of the six methods considered. For demonstration purposes, we have also included the multimodal sensing aspect discussed in \cref{ss:multimodal}. The magenta line indicates the trajectory of the agent through a graph of size $n=625$. The $m=20$ prediction locations are shown in white. The yellow points indicate where more accurate sensing locations were chosen. %
  For visualization purposes, the variance is displayed across the entire grid rather than at the prediction locations only.} \label{fig:trajectories}
\end{figure*}

The convex mixed integer problem for the multi-agent case is then: \begin{align}
\text{minimize} \quad & \sum_{j=1}^M \phi(z^j)  \nonumber \\
\text{subject to} \quad & z^j \in \mathcal{Z} \quad \forall j = 1, \dots, M
\end{align} which can be handled by our approximate sequential path optimization in a similar fashion as presented in \cref{sec:cocp}.

\section{Results}\label{sec:results}

We provide empirical evidence of the advantages of our approach over a variety of other baselines and methods from the literature. Namely, we directly compare our method with an implementation of the mixed integer formulation presented by \citeauthor{dutta2022informative} as well as the belief MDP formulation of \citeauthor{ott2023sequential} \cite{dutta2022informative, ott2023sequential}. Our experiments focus on the interplay between the number of nodes in the graph, the allotted budget, and the allowed runtime to evaluate the influence on the information objective of interest. All of the results presented in this section were run on a computer with an M1 Max and 64 GB of RAM.

\subsection{Experimental Setup}
We consider environments modeled as grid-based graphs, but our approach scales to graphs with arbitrary structures. The graph is a $\sqrt{n} \times \sqrt{n}$ grid contained in an environment with constant edge lengths. %
Our experimental setup is similar to that presented by \citeauthor{dutta2022informative}; to allow for consistent comparisons we use a prediction set fixed at $m=20$ \cite{dutta2022informative}. In all experiments, the start and end nodes are on diagonally opposite ends of the grid. The covariance structure of the environment is modeled using the squared exponential kernel \begin{equation}k(x, x') = \exp \left(- \frac{(x-x')^2}{2 \ell^2 } \right) \label{eq:kernel_function}\end{equation} where $\ell = 1.0$ is the length scale. Using this covariance structure, the prior density for $x$ is given by $\mathcal{N}(0, \Sigma_x)$ where  \begin{equation}
    \Sigma_x = \begin{bmatrix}
    k(x_1, x_1) & \dots & k(x_1, x_n) \\
    \vdots & \ddots & \vdots \\
    k(x_n, x_1) & \dots & k(x_n, x_n)
\end{bmatrix}.
\end{equation}
In all of the following experiments, we limit the maximum runtime for each method to $120$ seconds to ensure even comparison across all approaches considered.

\begin{figure}[t]
\centering
    {\includegraphics[width=1\columnwidth]{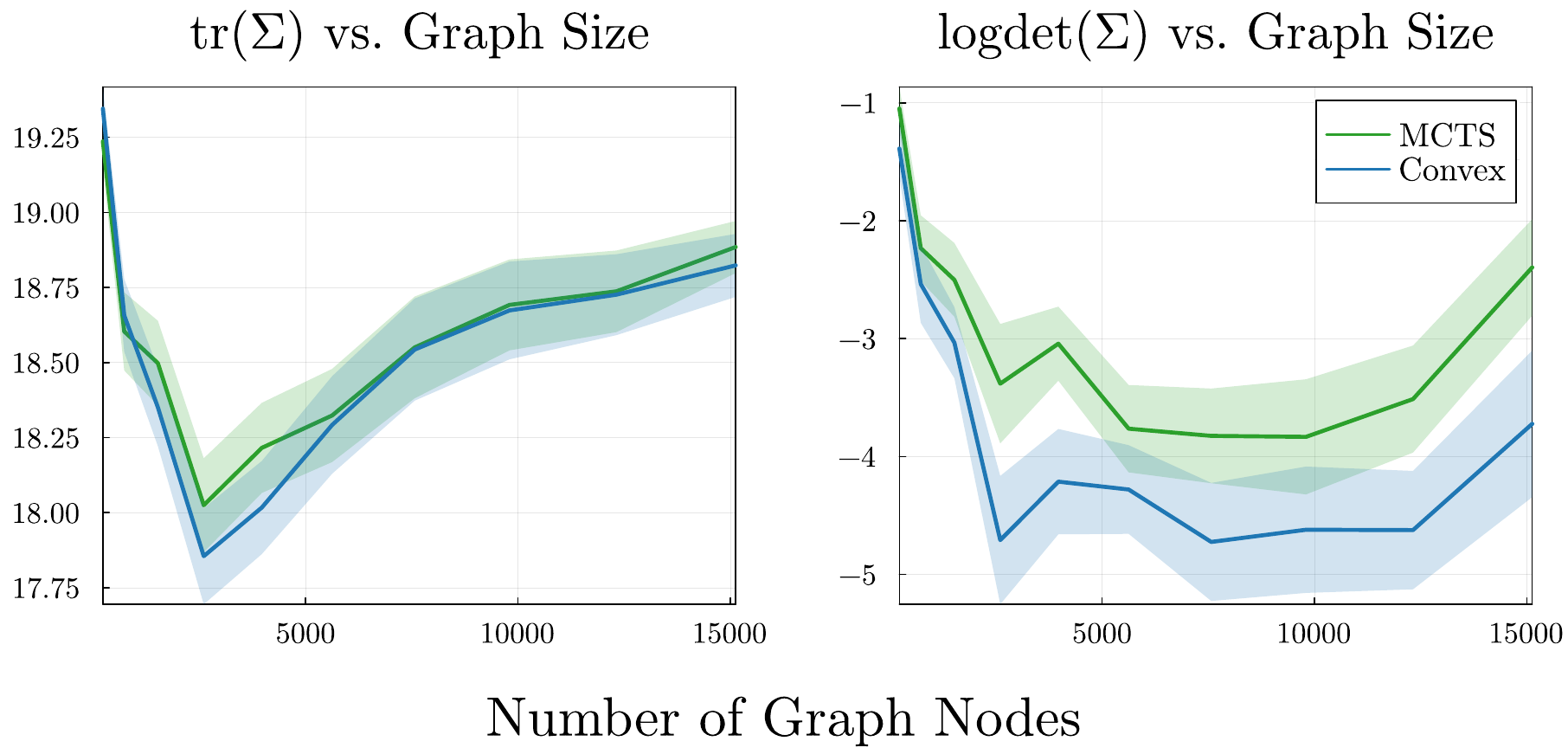}}    
  \caption{Comparison of the convex multimodal sensor selection against MCTS multimodal sensor selection. These comparisons were done with $k=3$ high-quality sensors. We can see that the convex multimodal sensor selection tends to select more informative distributions of the high-quality sensors for both the A and D-IPP objectives.} \label{fig:mcts_cvx_drill}
\end{figure}

\subsection{B-IPP as a Surrogate Objective for A-IPP \label{ss:approx_a_opt_objective}} %

We first begin with the A-IPP problem and compare the exact mixed integer convex problem (MICP) presented in \cref{eq:exact_ipp} with the exact MICP using the B-IPP $\mathbf{tr}(\Sigma^{-1})$ objective along with the mixed integer program (MIP) formulation from \citeauthor{dutta2022informative} \cite{dutta2022informative}. For the exact MICP using the B-IPP objective we are optimizing the path with respect to the B-IPP objective and then evaluating the path with respect to the A-IPP objective. Note that the MIP formulation presented by \citeauthor{dutta2022informative} only considers the A-IPP problem and does not provide an equivalent formulation for the D-IPP objective. We first compare our results on small graphs ranging from $n=4$ to $n=3844$. Our results show that on small graph sizes ($n<500$), all three of the methods perform similarly in both runtime and solution quality as shown in \cref{fig:mip_exact_tr_comparison}. %

However, when we scale up to larger graph sizes, we can see that the $\mathbf{tr}(\Sigma^{-1})$ objective can scale much more efficiently. Note that all methods are restricted to the same $120$ second time constraint and the best solution found is then returned. The MIP approach performs well for $n<500$ but fails to scale to larger graphs. Most notably, the B-IPP objective proves to be an effective surrogate for the A-IPP problem. The reason for the improved performance is twofold. First, the $\mathbf{tr}(\Sigma^{-1})$ objective is an affine function of the decision variable $z$ resulting in a mixed integer linear program. Second, by maximizing $\mathbf{tr}(\Sigma^{-1})$, we are trying to increase the magnitude of the eigenvalues of \(\Sigma_z^{-1}\). We know that the eigenvalues of \(\Sigma_z^{-1}\) are the reciprocals of the eigenvalues of \(\Sigma_z\) (assuming \(\Sigma_z\) is invertible). Therefore, increasing the eigenvalues of \(\Sigma_z^{-1}\) will decrease the corresponding eigenvalues of \(\Sigma_z\). While maximizing \(\mathbf{tr}(\Sigma_z^{-1})\) indirectly influences the eigenvalues of \(\Sigma_z\), the trace of \(\Sigma_z\) and the trace of \(\Sigma_z^{-1}\) are not simply reciprocals of each other. %
That is, the A and B-IPP objectives are not equivalent, but they are closely related and a meaningful performance improvement can be achieved using the B-IPP objective as a surrogate on larger problem instances.
The goal of these experiments was to isolate the comparison between these three approaches to the A-IPP objective. We now proceed to scale our experiments to even larger graphs and include both A and D-IPP objectives with additional baselines to compare against.

\subsection{A and D-IPP Results}
We compare the following approaches on both the A and D-IPP objectives:
\begin{enumerate}
    \item \textbf{Exact}: the solution obtained from solving the exact MICP presented in \cref{eq:exact_ipp}.
    \item{$\textbf{tr}(\Sigma^{-1})$: the solution obtained from solving the exact MICP using the B-IPP objective. 
    \item \textbf{MCTS}: the belief MDP approach presented by \citeauthor{ott2023sequential} \cite{ott2023sequential}.
    \item \textbf{Random}: starting from the initial vertex, the agent chooses randomly from the neighboring edges to then transition to the next vertex. The action space is constrained so that the agent can always reach the goal location within the specified budget constraint.
    \item \textbf{Greedy}: starting from the initial vertex, the agent evaluates the improvement in the objective from taking each of the neighboring edges. The agent then selects the edge that results in the greatest performance and repeats this process. The action space is constrained so that the agent must always reach the goal location within the specified budget constraint.
    \item{\textbf{ASPO}: the approximate sequential path optimization approach from \cref{sec:cocp}.}}
\end{enumerate}

\begin{figure}[t]
\centering
    {\includegraphics[width=1\columnwidth]{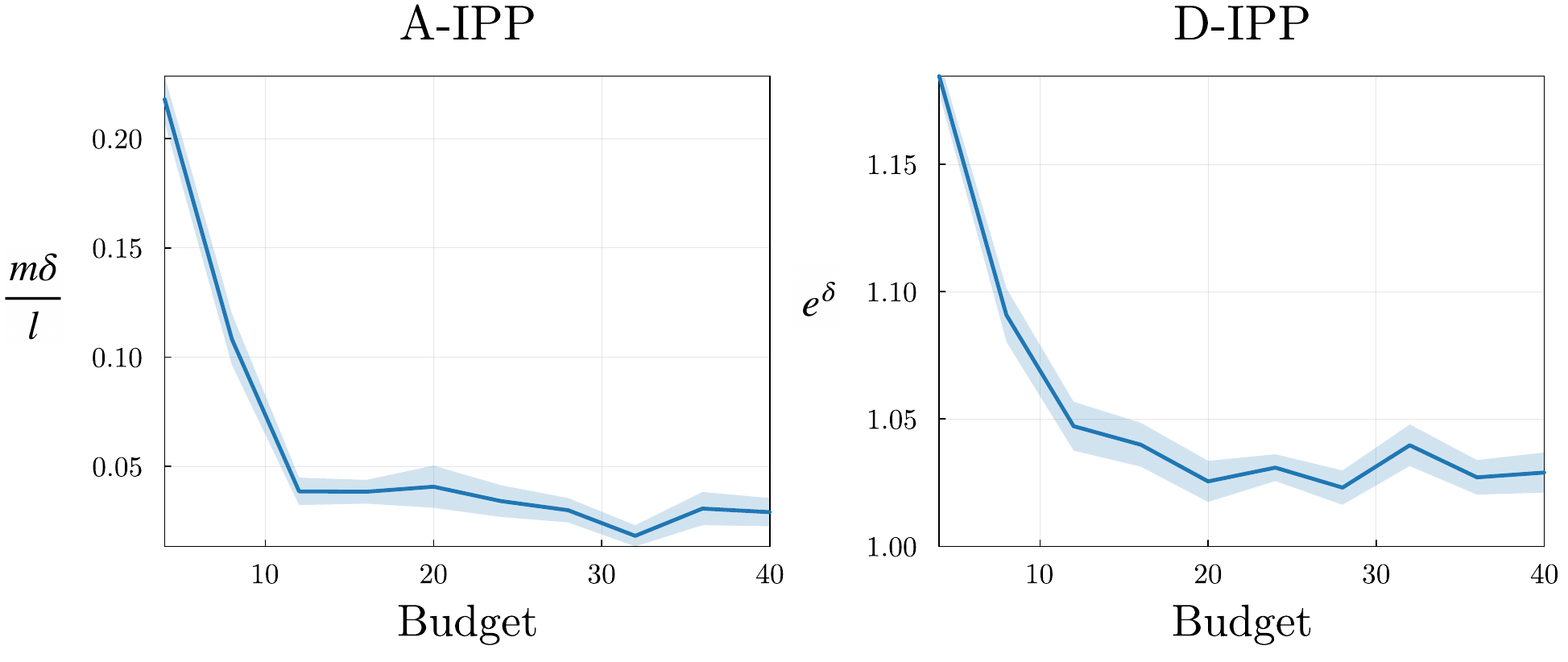}}    
  \caption{Results showing the average optimality gap and standard error as the budget is increased for the ASPO method with $N=1600$. Here the gap is reported normalized by the lower bound for A-IPP and exponentiated for D-IPP.} %
  \label{fig:gap}
\end{figure}

\begin{figure*}[t]
\centering
    {\includegraphics[width=1\textwidth]{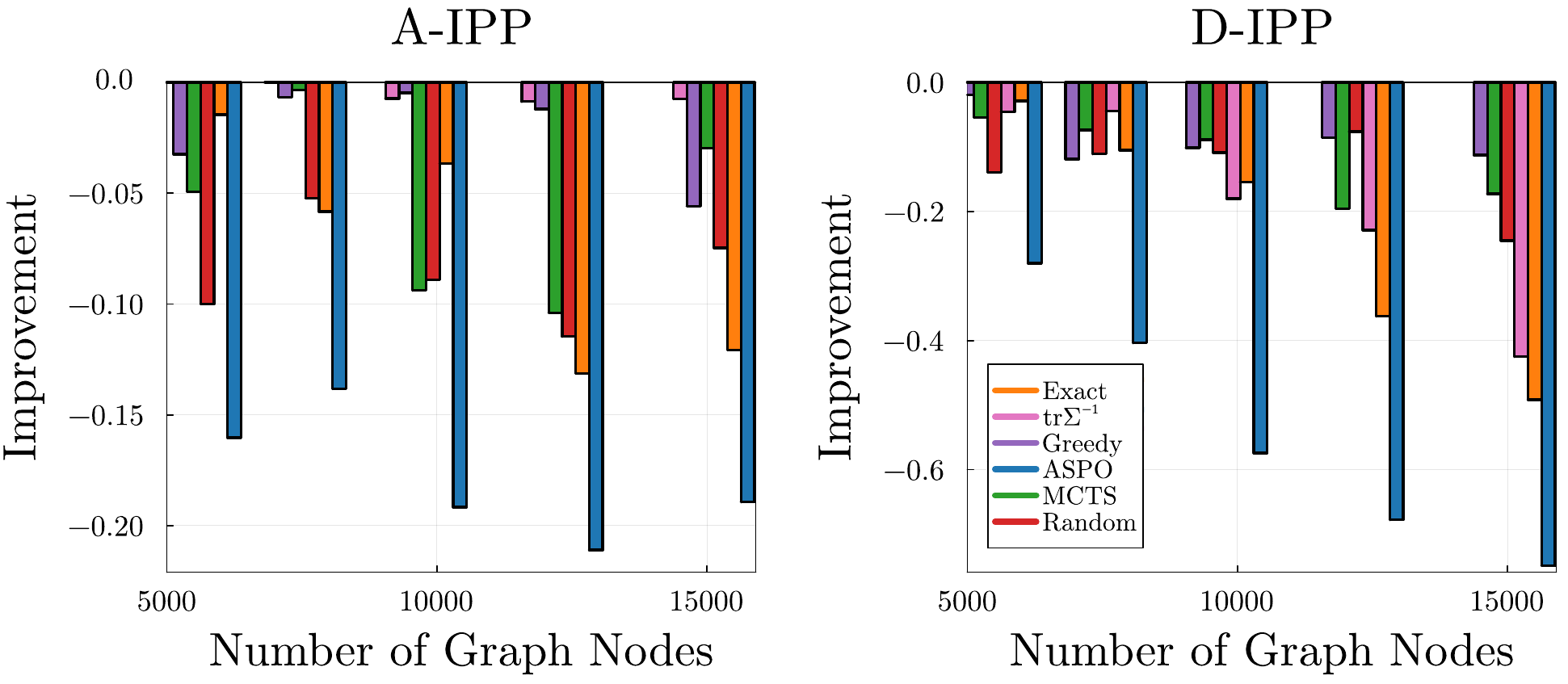}}    
  \caption{Results from the local optimization procedure. The average improvement given by the reduction in objective value over 25 simulation runs is shown.} \label{fig:local_optimization}
\end{figure*}

\Cref{fig:a-optimal} shows the runtime and objective value as a function of graph size for the A-IPP objective evaluated on each of the six methods discussed above. Note that we used a modified version of MCTS where the number of iterations performed in the tree search was scaled proportional to the runtime budget used so far. The reason for doing this was to allow for better performance of MCTS by allowing for a dynamic adjustment of the iterations so that the entire $120$ second runtime constraint would be used. As a result, we see that MCTS uses the full runtime constraint even for relatively small graphs. MCTS performs competitively with graphs of up to $n=1000$ at which point the performance begins to degrade as the runtime constraint leads to fewer iterations since more individual action steps are required to make it to the goal location. We see that the $\textbf{tr}(\Sigma^{-1})$ objective is the best performer on graph sizes of $n<5000$ and then is outperformed by our ASPO method. ASPO performs similarly to the $\textbf{tr}(\Sigma^{-1})$ objective on smaller graphs and outperforms all of the methods considered on larger graphs while also reducing the runtime by more than a factor of two relative to the $\textbf{tr}(\Sigma^{-1})$ objective.

\Cref{fig:d-optimal} shows the runtime and objective value as a function of graph size for the D-IPP objective evaluated on each of the six methods. Again, the $\textbf{tr}(\Sigma^{-1})$ objective proves to perform quite well even when evaluated on the D-IPP objective. In this case, the exact and $\textbf{tr}(\Sigma^{-1})$ methods perform similarly. MCTS exhibits similar performance to the A-IPP case where it performs competitively on small graph sizes and then degrades on larger graph sizes due to the runtime constraints. ASPO outperforms both the exact and $\textbf{tr}(\Sigma^{-1})$ methods while again reducing runtime by more than a factor of two and scaling performance to the largest graph sizes considered.

Examples of the trajectories produced by each of the methods are shown in \cref{fig:trajectories}. In these example trajectories, we have also included the multimodal sensing aspect of the problem introduced in \cref{ss:multimodal}. MCTS used the method described by \citeauthor{ott2023sequential} to select locations for the high-quality sensors \cite{ott2023sequential}. The random method selected high-quality sensor locations from its action space and was limited to placing $k=3$ high-quality sensors. The other methods all used the convex formulation provided in \cref{ss:multimodal}. From the trajectories, we can see that the random solution is the worst performer and stays near the starting location (bottom left corner) until it reaches the budget limit and must head to the goal location leaving much of the environment unexplored. The greedy method slightly improves upon the random method, but is unable to coordinate its actions several steps into the future therefore leading to inefficient exploration. MCTS improves upon the greedy solution; however, much of its exploration is limited to small areas around where it has already visited. The exact and $\textbf{tr}(\Sigma^{-1})$ methods are able to better distribute the measurement locations resulting in more efficient exploration. Lastly, the approximate sequential path optimization method tries to fully cover the environment by distributing its measurements accordingly and outperforms the exact and $\textbf{tr}(\Sigma^{-1})$ methods in terms of the best (lowest) value of the A-IPP objective.

\begin{figure*}[t]
\centering
    {\includegraphics[width=1\textwidth]{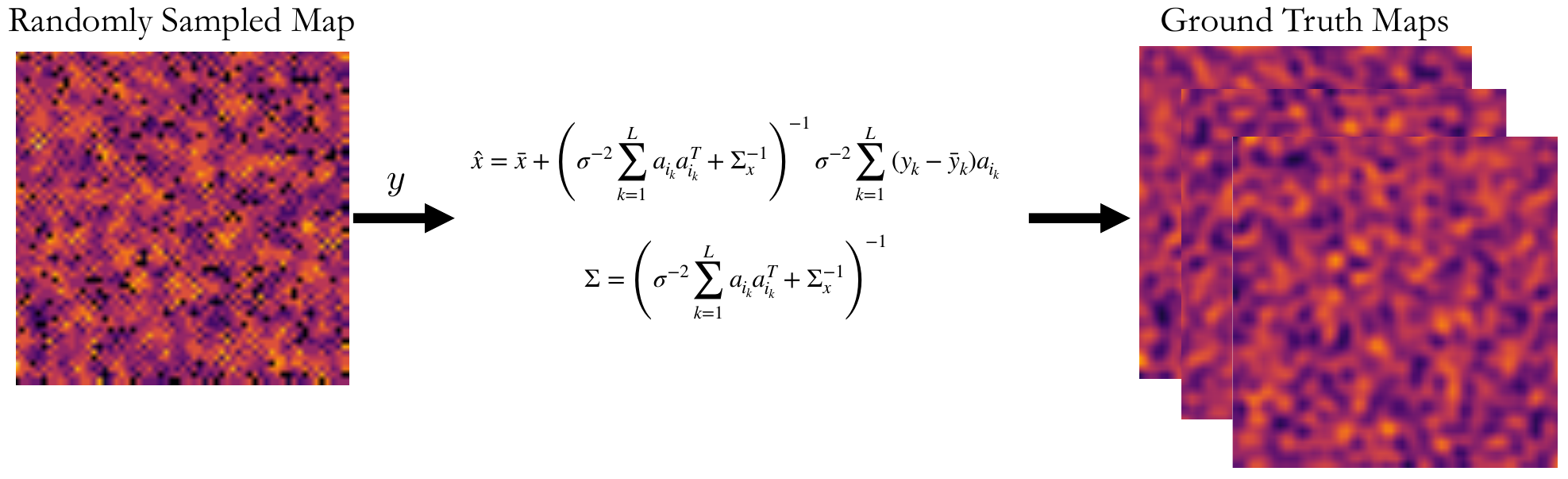}}    
  \caption{Illustration showing the construction of the ground truth maps used in our adaptive IPP experiments. Note the length scale in \cref{eq:kernel_function} controls the amount of spatial correlation in the environment.} \label{fig:ground_truth_map_generation}
\end{figure*}

\subsection{Quality of Convex Sensor Selection vs. MCTS}
\Cref{fig:mcts_cvx_drill} compares the multimodal sensor selection methods of \citeauthor{ott2023sequential} and the method we present in \cref{ss:multimodal}. For these experiments, we started with the paths that were produced from the MCTS policy. To construct these paths, the MCTS policy chooses from an action space that includes both movement and sensing actions. Therefore, each point along the trajectory includes a location and sensor type to be used. To compare the sensor selection methods, we left the MCTS path solution constant and used this as the initial path for the convex multimodal sensor selection problem. We then evaluated both sensor selection methods on the A and D-IPP objectives (i.e. the paths were fixed for both methods, but the sensor types were changed). \Cref{fig:mcts_cvx_drill} shows that the convex multimodal sensor selection method we propose outperforms that of MCTS. Note that these comparisons were done with $k=3$ high-quality sensors.

\subsection{Optimality Gap \label{ss:results_optimality_gap}}

\Cref{fig:gap} shows the empirical results of the optimality gap using our approximate sequential path optimization method as the budget constraint is varied. We can see that as the budget is increased, our solution decreases the optimality gap. Here, $\frac{m \delta}{l}$ quantifies how close the solution is to the lower bound for the A-IPP objective. The factor of $m$ is included for appropriate scaling with the definition of $\delta$ given in \cref{eq:delta}. For the D-IPP objective, we use $e^{\delta}$ which is the ratio of mean radii of the confidence ellipsoid \cite{joshi2008sensor}. Additionally, we can see that our proposed method is always within at least $25 \%$ of the lower bound. In reality, we are often much closer to the true optimal value since the lower bound is not necessarily the true optimal value. 

\subsection{Local Optimization}
As discussed by \citeauthor{joshi2008sensor} and \citeauthor{diamond2018general}, we can perform a local optimization step to greedily improve our approximate solution \cite{joshi2008sensor, diamond2018general}. This local optimization procedure is the polish step in the relax-round-polish method presented by \citeauthor{diamond2018general} \cite{diamond2018general}. Starting from the approximate solution $\hat{z}$ presented in \cref{sec:cocp}, we check other paths that can be derived from $\hat{z}$ by swapping one of the $k$ nodes along the path with one of the one-hop neighbors available. In this case, we define a valid one-hop neighbor of node $i$ to be a node $j$ such that $j \in N_{out}(i)$ and if node $i$ were removed from the path and node $j$ were added then the total length of the path would not change. Equivalently, if $l_A = \sum_{i=1}^n \sum_{j \in N_{out}(i)} \hat{z}_{ij}$ represents the total length of path $A$ before a valid swap occurs then $l_A$ must equal $l_B$ where path $B$ contains a valid swap derived from path $A$. 

Given the current approximate solution $\hat{z}$, we randomly sample a node $j$ from the path and check to see if any of its valid one-hop neighbors improve the objective when swapped with node $j$. If so, we swap the nodes and continue the random search. In theory, we could search over all possible one-hop neighbors of the path and continue this search until no swap is found to improve the objective. The resulting solution would be the $2-opt$ solution because exchanging any node on the path with any valid one-hop neighbor will not improve the solution. Doing so can involve a large number of local optimization steps, so in practice, we can simply limit the number of steps taken to $N_{\text{loc}}$. 

\begin{figure*}[t]
\centering
    {\includegraphics[width=1\textwidth]{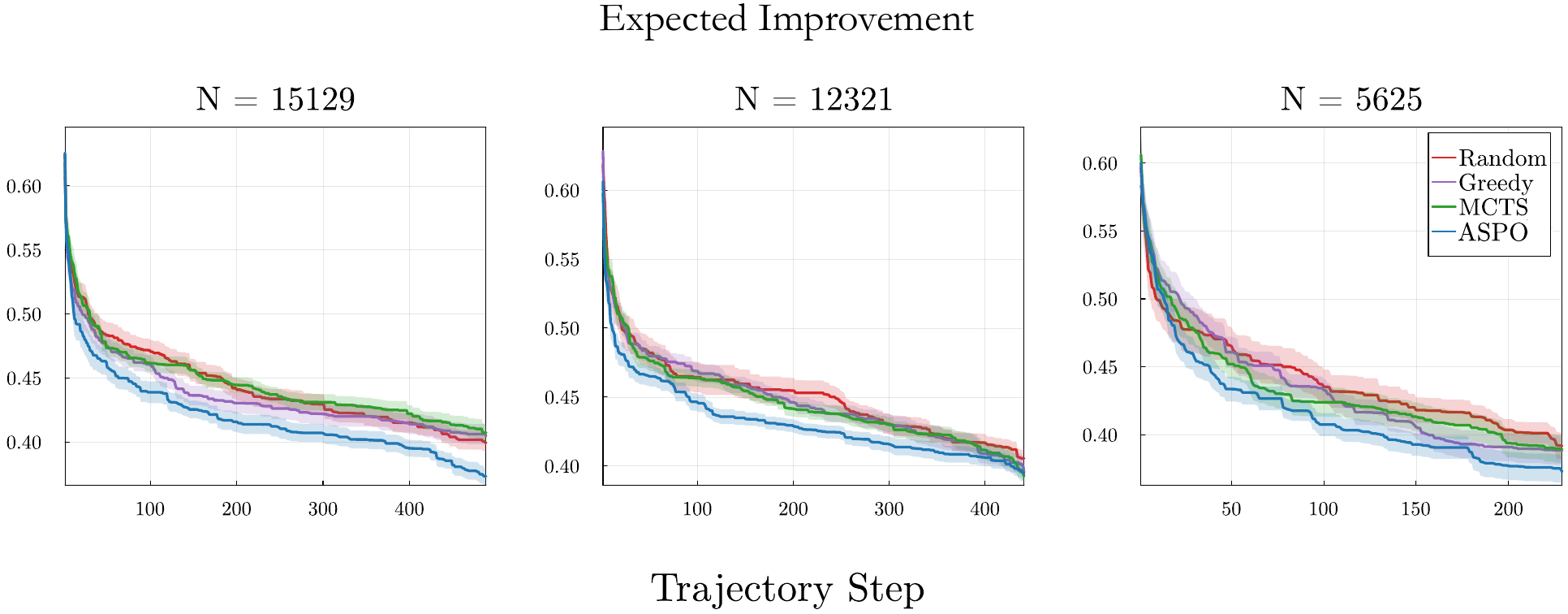}}    
  \caption{Results showing the net expected improvement normalized by graph size as the trajectory progresses for each of the methods that support adaptive objectives.} \label{fig:expected_improvement}
\end{figure*}

The results from this local optimization polishing procedure are shown in \cref{fig:local_optimization}. We can see that the local optimization tends to improve the performance of each of the methods. In some cases, we see that the local optimization has a greater effect than others. Namely, it has a consistently larger effect on the exact and ASPO methods for both the A and D-IPP objectives. The $\textbf{tr}(\Sigma^{-1})$ method benefits from an improvement on the D-IPP objective, while only receiving a small improvement on the A-IPP objective. Typically, methods that start with reasonable solutions to begin with, such as ASPO and the exact method, can achieve a significant performance boost by using the simple local optimization procedure since only valid one-hop swaps are considered. However, this is not always the case as indicated by the difference in improvement for the  $\textbf{tr}(\Sigma^{-1})$ method between A and D-IPP objectives.

\subsection{Other Objectives}
As discussed in \cref{ss:additional_objectives}, our ASPO method can handle the adaptive version of the IPP problem where the objective depends on the received sample values. The previous experiments were not concerned with the actual sensor observation values received at each of the nodes in the environment, but in the adaptive IPP problem, these sensor observations now become part of the adaptive objective.

For these experiments, we construct a ground truth spatial map of the environment containing the true quantity of interest. The environment is modeled as an $\sqrt{n} \times \sqrt{n}$ grid where each cell contains a value for the true quantity of interest. To construct a spatially-correlated environment, we first start by randomly generating a value for each grid cell. We then use these randomly generated values as the $y$ values in \cref{eq:gp_mean} using the covariance structure from \cref{eq:kernel_function}. This gives us the mean and covariance of the Gaussian process which we use to sample different ground truth maps for different simulation runs. An illustration of this process is shown in \cref{fig:ground_truth_map_generation}.

\Cref{fig:expected_improvement} shows our experiment results for the adaptive IPP problem. For this set of experiments, we only compared the Random, Greedy, MCTS, and ASPO methods since these were the only methods that supported adaptive objectives. As we can see, ASPO outperforms the other methods in all graph sizes considered and consistently results in a net expected improvement remaining in the environment of $0.4$ or less when normalized by graph size. As expected, we see the random method tends to perform the worst. In large graphs, we see that MCTS and Greedy only perform marginally better due to the significant increase in the size of the search space.

These results empirically support the teleportation approximation used by the ASPO method to estimate the value of visiting different regions of the environment. In addition, it demonstrates the flexibility of the ASPO method to support both adaptive and non-adaptive objectives, as well as incorporating multimodal sensing. In this way, the ASPO method can outperform the other methods considered on the IPP, AIPP, and AIPPMS problems considered.

\begin{figure*}[t]
\centering
    {\includegraphics[width=1.0\textwidth]{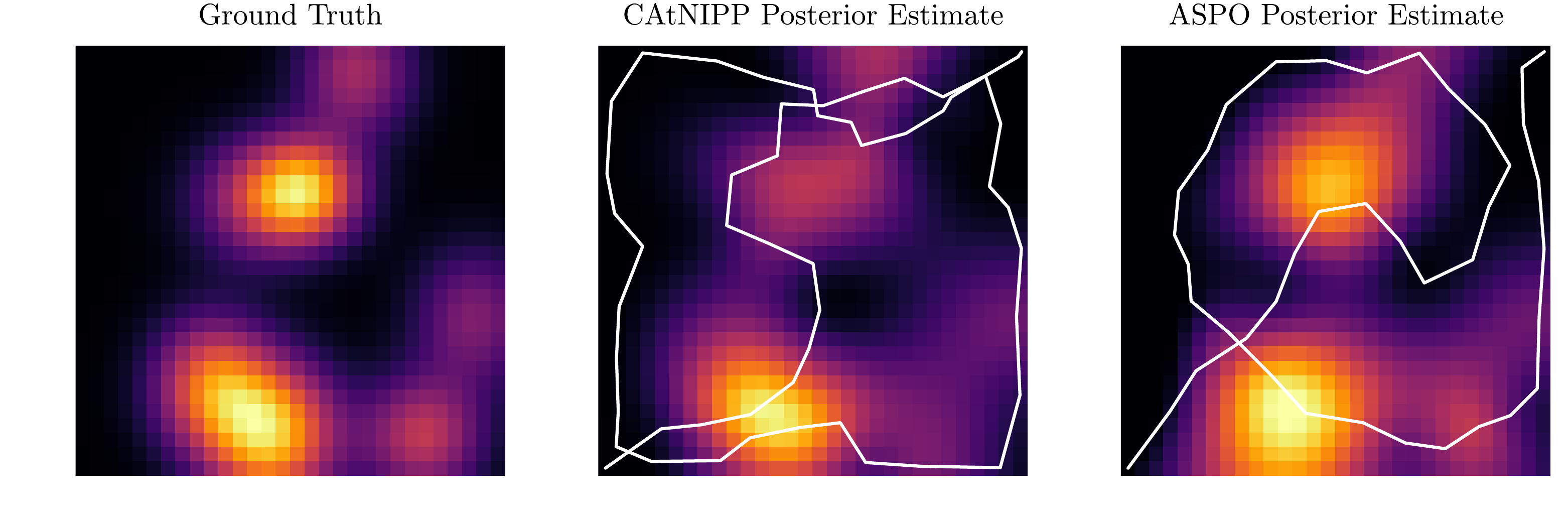}}    
  \caption{\textcolor{black}{Examples of trajectories planned by the CAtNIPP trajectory sampling variant and the ASPO method on a graph with $400$ nodes and available budget of $6$. }} \label{fig:catnipp_aspo}
\end{figure*}

\begin{table*}[h]
\textcolor{black}{
\centering
\footnotesize %
\setlength{\tabcolsep}{4pt} %
\begin{tabular}{l@{\hskip 8pt}l@{\hskip 8pt}l@{\hskip 16pt}l@{\hskip 8pt}l@{\hskip 16pt}l@{\hskip 8pt}l@{\hskip 16pt}l@{\hskip 8pt}l@{\hskip 16pt}l@{\hskip 8pt}l}
\toprule
\multirow{2}{*}{Method} & \multicolumn{2}{c}{Budget 4} & \multicolumn{2}{c}{Budget 6} & \multicolumn{2}{c}{Budget 8} & \multicolumn{2}{c}{Budget 10} & \multicolumn{2}{c}{Budget 12} \\
\cmidrule(lr){2-3} \cmidrule(lr){4-5} \cmidrule(lr){6-7} \cmidrule(lr){8-9} \cmidrule(lr){10-11}
& $\mathbf{tr}(\Sigma)$ & T(s) & $\mathbf{tr}(\Sigma)$ & T(s) & $\mathbf{tr}(\Sigma)$ & T(s) & $\mathbf{tr}(\Sigma)$ & T(s) & $\mathbf{tr}(\Sigma)$ & T(s) \\
\midrule
CG & 19.67($\pm$1.57) & \textbf{0.15} & 5.42($\pm$0.29) & \textbf{0.19} & 2.57($\pm$0.14) & \textbf{0.26} & 1.58($\pm$0.09) & \textbf{0.33} & 1.24($\pm$0.07) & \textbf{0.42} \\
CTS & 15.94($\pm$1.18) & 44.04 & 4.94($\pm$0.23) & 62.35 & 2.27($\pm$0.12) & 88.78 & 1.54($\pm$0.08) & 111.88 & 1.24($\pm$0.07) & 143.94 \\
ASPO & \textbf{14.02($\pm$0.87)} & 0.69 & \textbf{4.79($\pm$0.28)} & 1.26 & \textbf{2.24($\pm$0.13)} & 1.96 & \textbf{1.28($\pm$0.11)} & 2.61 & \textbf{0.98($\pm$0.11)} & 3.33 \\
\bottomrule
\end{tabular}}
\caption{\textcolor{black}{Comparison between CAtNIPP Greedy (CG), CAtNIPP Trajectory Sampling (CTS), and ASPO IPP solvers on the AIPP problem using the environment setup presented by \citeauthor{cao2022catnipp} with $400$ nodes, $20$ edge connections, and $4$ trajectory samples for CTS \cite{cao2022catnipp}. The average covariance matrix trace $\mathbf{tr}(\Sigma)$ in high interest areas is reported after running out of budget (standard error in parentheses). T(s) is the average total planning time in seconds. Results from $100$ simulation runs for each budget.}}  \label{table:catnipp_table}
\end{table*}

\subsection{Comparison with Attention-based Network Approach}
\textcolor{black}{We also compared our approximate method with the Context-Aware Attention-based Network for Informative Path Planning (CAtNIPP) approach presented by \citeauthor{cao2022catnipp} \cite{cao2022catnipp}. CAtNIPP is a state-of-the-art deep reinforcement learning based approach using attention-based neural networks to learn policies that balance short-term exploitation and long-term exploration for the AIPP problem.} 

\textcolor{black}{For even comparison between our two methods, we used the graph and environment setup described by \citeauthor{cao2022catnipp} \cite{cao2022catnipp}. Specifically, a Matérn $3/2$ Kernel with length scale of $0.45$. The ground truth environment is constructed by averaging $8$ to $12$ random 2-dimensional Gaussian distributions in the unit square $[0, 1]^2$, to construct the true interest map \cite{cao2022catnipp}.}

\textcolor{black}{The objective is to reduce the variance in high-interest areas defined as regions that have true quantities of interest greater than or equal to $0.4$. To effectively solve this variant of the AIPP problem, solution methods must balance the tradeoff between collecting additional measurements in high-interest areas they have previously discovered with exploring new areas of the environment they have not yet visited. \citeauthor{cao2022catnipp} present two different variants of CAtNIPP. The greedy variant works in the standard reinforcement learning manner where the agent selects the action with the highest activation in its policy. The trajectory sampling variants plan a $15$-step trajectory and execute the first $3$ steps before replanning.}

\textcolor{black}{\Cref{table:catnipp_table} presents the results across several different budgets. We see that across all of the budgets, our ASPO method consistently achieves the lowest variance in high-interest areas showing that it more effectively balances the exploration exploitation tradeoff. We see that the CAtNIPP Greedy variant has the lowest planning time across all of the methods considered but also is consistently the worst performer. ASPO is quite close to the Greedy variant in terms of planning time, while the trajectory sampling variant requires significantly more planning time. \Cref{fig:catnipp_aspo} also shows examples of trajectories planned by the CAtNIPP trajectory sampling variant and ASPO in addition to the posterior estimates after executing the trajectories.}

\textcolor{black}{It is important to note that this comparison was performed on the specific environment that CAtNIPP was trained on. CAtNIPP is sensitive to changes in the hyperparameters of the Gaussian process kernel function and the upper bound confidence that defines the high-interest areas. Therefore, CAtNIPP would require retraining if any of these parameters were changed significantly. \citeauthor{cao2022catnipp} report that it took around $24$ hours training on four GPUs to converge. Therefore, one of the key strengths of ASPO over neural network based approaches is that no retraining is required when any of the Gaussian process hyperparameters, graph structures, budgets, objectives, or confidence intervals are changed.}

\subsection{Multi-agent Example}

\begin{figure*}[t]
\centering
    {\includegraphics[width=0.8\textwidth]{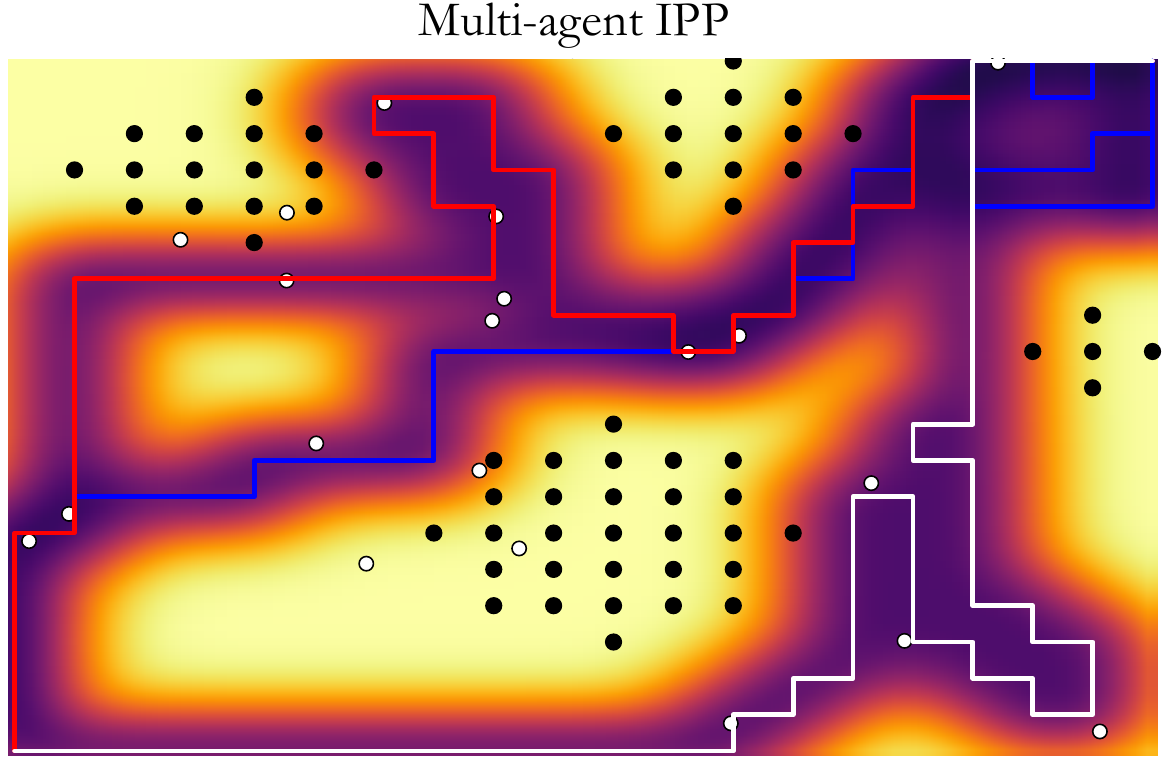}}    
  \caption{Example of the multi-agent IPP solution produced with our ASPO method. Obstacles in the environment are shown by black circles. The $m=20$ prediction locations are shown by white circles. The three agent trajectories are shown in red, blue, and white.} \label{fig:multi-agent}
\end{figure*}

\textcolor{black}{\Cref{fig:multi-agent} illustrates the application of the ASPO method in a multi-agent Informative Path Planning (IPP) scenario, where \( M=3 \) agents are deployed in an environment that includes obstacles. Each agent \( A_i \) is assigned the task of traversing through the environment to collect measurements. The measurements gathered by the individual agents are then shared among them, resulting in a collective measurement vector that represents the combined data from all agents.}

\textcolor{black}{We can see that the agents are able to distribute their exploration throughout the environment to effectively split up the environment into thirds leading to more efficient exploration. %
Notably, the solution to this multi-agent problem, computed using our ASPO method, was achieved in less than $5$ seconds, on a graph with $400$ nodes. This highlights the efficiency and scalability of the ASPO method, particularly in complex environments with multiple agents and obstacles.}

\section{Conclusion}\label{sec:conclusion}

This study focused on planning informative paths in graphs using information-theoretic objectives. We demonstrated the challenge that exact solution methods such as mixed integer programs have in scaling to larger graph sizes. We introduced the B-IPP objective and showed that its formulation as a mixed integer linear program positions it as an effective surrogate objective for both the A and D-IPP problems. We proposed a novel approximate sequential path optimization method using dynamic programming to produce approximate solutions to the IPP problem. Our ASPO method achieves a balance between computational feasibility and solution quality. By leveraging the strengths of dynamic programming, our method iteratively constructs paths, incorporating the updated information distribution at each step.

We provided a bound on the optimality of our approximate method, confirming that our approximations are not only efficient but also consistently near optimal. Furthermore, the flexibility of our method is evident in its capability to handle adaptive objectives, multimodal sensing, and multi-agent extensions of the classic IPP problem. This flexibility highlights the capability of our method in the identification and prioritization of pertinent regions based on received measurement values. We demonstrated the efficacy of our ASPO method in larger graphs showing that solutions could be produced on graphs with more than 4,000 nodes and 15,000 edges within seconds. The flexibility and efficiency of our method are crucial for real-world settings in dynamic environments where the relevance of information can change rapidly. 

\subsection{Limitations and Future Work}
\textcolor{black}{While our method has proven effective, there are some key limitations that present opportunities for future research. First, our approach assumes that the location of the agent is fully observable, the current energy state of the agent is known, and the amount of budget used by specific actions is deterministic. Incorporating uncertainty in these aspects would make our method more robust and better aligned with real-world scenarios where localization errors, energy consumption variability, and unexpected environmental changes are common. Additionally, our current work assumes full knowledge of these variables, and introducing uncertainty in these areas could lead to more sophisticated and resilient planning strategies.}

\textcolor{black}{Another promising area for future research involves exploring more complex multi-agent scenarios. Specifically, we aim to investigate cases where communication between agents is limited or unreliable, which is a common challenge in real-world applications. Additionally, introducing adversarial agents into the problem space, which may attempt to disrupt or compete with the planning agents, could add another layer of complexity and realism to the problem. Addressing these challenges will likely require the development of new approximate methods that can maintain scalability while handling the increased combinatorial complexity introduced by these factors.}

\section*{Acknowledgments}\label{sec:acknowledgements}

We would like to thank Shamak Dutta for his insightful discussions and support in comparing with his previous work. We would also like to thank Lauren Ho-Tseung for her contributions and support in developing this method.

\newpage
\bibliographystyle{IEEEtranN}
\bibliography{references}

\begin{thebibliography}{31}
\providecommand{\natexlab}[1]{#1}
\providecommand{\url}[1]{#1}
\csname url@samestyle\endcsname
\providecommand{\newblock}{\relax}
\providecommand{\bibinfo}[2]{#2}
\providecommand{\BIBentrySTDinterwordspacing}{\spaceskip=0pt\relax}
\providecommand{\BIBentryALTinterwordstretchfactor}{4}
\providecommand{\BIBentryALTinterwordspacing}{\spaceskip=\fontdimen2\font plus
\BIBentryALTinterwordstretchfactor\fontdimen3\font minus \fontdimen4\font\relax}
\providecommand{\BIBforeignlanguage}[2]{{%
\expandafter\ifx\csname l@#1\endcsname\relax
\typeout{** WARNING: IEEEtranN.bst: No hyphenation pattern has been}%
\typeout{** loaded for the language `#1'. Using the pattern for}%
\typeout{** the default language instead.}%
\else
\language=\csname l@#1\endcsname
\fi
#2}}
\providecommand{\BIBdecl}{\relax}
\BIBdecl

\bibitem[Joshi and Boyd(2008)]{joshi2008sensor}
S.~Joshi and S.~Boyd, ``Sensor selection via convex optimization,'' \emph{IEEE Transactions on Signal Processing}, vol.~57, no.~2, pp. 451--462, 2008.

\bibitem[Dutta et~al.(2022)Dutta, Wilde, and Smith]{dutta2022informative}
S.~Dutta, N.~Wilde, and S.~L. Smith, ``Informative path planning in random fields via mixed integer programming,'' in \emph{IEEE Conference on Decision and Control (CDC)}.\hskip 1em plus 0.5em minus 0.4em\relax IEEE, 2022, pp. 7222--7228.

\bibitem[Meliou et~al.(2007)Meliou, Krause, Guestrin, and Hellerstein]{meliou2007nonmyopic}
A.~Meliou, A.~Krause, C.~Guestrin, and J.~M. Hellerstein, ``Nonmyopic informative path planning in spatio-temporal models,'' in \emph{AAAI Conference on Artificial Intelligence (AAAI)}, vol.~10, no.~4, 2007, pp. 16--7.

\bibitem[Morere et~al.(2017)Morere, Marchant, and Ramos]{morere2017sequential}
P.~Morere, R.~Marchant, and F.~Ramos, ``Sequential {B}ayesian optimization as a {POMDP} for environment monitoring with {UAV}s,'' in \emph{IEEE International Conference on Robotics and Automation (ICRA)}, 2017, pp. 6381--6388.

\bibitem[Popovi{\'c} et~al.(2020)Popovi{\'c}, Vidal-Calleja, Hitz, Chung, Sa, Siegwart, and Nieto]{popovic2020informative}
M.~Popovi{\'c}, T.~Vidal-Calleja, G.~Hitz, J.~J. Chung, I.~Sa, R.~Siegwart, and J.~Nieto, ``An informative path planning framework for {UAV}-based terrain monitoring,'' \emph{Autonomous Robots}, vol.~44, no.~6, pp. 889--911, 2020.

\bibitem[Vashisth et~al.(2024)Vashisth, Ruckin, Magistri, Stachniss, and Popovic]{vashisth2024deep}
A.~Vashisth, J.~Ruckin, F.~Magistri, C.~Stachniss, and M.~Popovic, ``Deep reinforcement learning with dynamic graphs for adaptive informative path planning,'' \emph{IEEE Robotics and Automation Letters}, 2024.

\bibitem[Yu et~al.(2022)Yu, Deng, Gui, Zhu, and Yao]{yu2022efficient}
T.~Yu, B.~Deng, J.~Gui, X.~Zhu, and W.~Yao, ``Efficient informative path planning via normalized utility in unknown environments exploration,'' \emph{Sensors}, vol.~22, no.~21, p. 8429, 2022.

\bibitem[Jakkala and Akella(2023)]{jakkala2023multi}
K.~Jakkala and S.~Akella, ``Multi-robot informative path planning from regression with sparse gaussian processes,'' \emph{arXiv preprint arXiv:2309.07050}, 2023.

\bibitem[Krause et~al.(2008)Krause, Singh, and Guestrin]{krause2008near}
A.~Krause, A.~Singh, and C.~Guestrin, ``Near-optimal sensor placements in {G}aussian processes: Theory, efficient algorithms and empirical studies,'' \emph{Journal of Machine Learning Research}, vol.~9, no.~2, 2008.

\bibitem[Singh et~al.(2009)Singh, Krause, Guestrin, and Kaiser]{singh2009efficient}
A.~Singh, A.~Krause, C.~Guestrin, and W.~J. Kaiser, ``Efficient informative sensing using multiple robots,'' \emph{Journal of Artificial Intelligence Research}, vol.~34, pp. 707--755, 2009.

\bibitem[Strawser and Williams(2022)]{strawser2022motion}
D.~Strawser and B.~Williams, ``Motion planning under uncertainty with complex agents and environments via hybrid search,'' \emph{Journal of Artificial Intelligence Research}, vol.~75, pp. 1--81, 2022.

\bibitem[B{\"a}ckstr{\"o}m et~al.(2021)B{\"a}ckstr{\"o}m, Jonsson, and Ordyniak]{backstrom2021cost}
C.~B{\"a}ckstr{\"o}m, P.~Jonsson, and S.~Ordyniak, ``Cost-optimal planning, delete relaxation, approximability, and heuristics,'' \emph{Journal of Artificial Intelligence Research}, vol.~70, pp. 169--204, 2021.

\bibitem[Bostr{\"o}m-Rost et~al.(2018)Bostr{\"o}m-Rost, Axehill, and Hendeby]{bostrom2018global}
P.~Bostr{\"o}m-Rost, D.~Axehill, and G.~Hendeby, ``On global optimization for informative path planning,'' \emph{IEEE Control Systems Letters}, vol.~2, no.~4, pp. 833--838, 2018.

\bibitem[Song et~al.(2015)Song, Rodriguez, and Teodorescu]{song2015trajectory}
S.~Song, A.~Rodriguez, and M.~Teodorescu, ``Trajectory planning for autonomous nonholonomic vehicles for optimal monitoring of spatial phenomena,'' in \emph{International Conference on Unmanned Aircraft Systems}.\hskip 1em plus 0.5em minus 0.4em\relax IEEE, 2015, pp. 40--49.

\bibitem[Hitz et~al.(2017)Hitz, Galceran, Garneau, Pomerleau, and Siegwart]{hitz2017adaptive}
G.~Hitz, E.~Galceran, M.-{\`E}. Garneau, F.~Pomerleau, and R.~Siegwart, ``Adaptive continuous-space informative path planning for online environmental monitoring,'' \emph{Journal of Field Robotics}, vol.~34, no.~8, pp. 1427--1449, 2017.

\bibitem[Francis et~al.(2019)Francis, Ott, Marchant, and Ramos]{francis2019occupancy}
G.~Francis, L.~Ott, R.~Marchant, and F.~Ramos, ``Occupancy map building through {B}ayesian exploration,'' \emph{The International Journal of Robotics Research}, vol.~38, no.~7, pp. 769--792, 2019.

\bibitem[Asgharivaskasi et~al.(2022)Asgharivaskasi, Koga, and Atanasov]{asgharivaskasi2022active}
A.~Asgharivaskasi, S.~Koga, and N.~Atanasov, ``Active mapping via gradient ascent optimization of {S}hannon mutual information over continuous {SE} (3) trajectories,'' in \emph{IEEE/RSJ International Conference on Intelligent Robots and Systems (IROS)}.\hskip 1em plus 0.5em minus 0.4em\relax IEEE, 2022, pp. 12\,994--13\,001.

\bibitem[Marchant et~al.(2014)Marchant, Ramos, Sanner, et~al.]{marchant2014sequential}
R.~Marchant, F.~Ramos, S.~Sanner \emph{et~al.}, ``Sequential {B}ayesian optimisation for spatial-temporal monitoring,'' in \emph{Conference on Uncertainty in Artificial Intelligence (UAI)}, 2014, pp. 553--562.

\bibitem[Morere et~al.(2018)Morere, Marchant, and Ramos]{morere2018continuous}
P.~Morere, R.~Marchant, and F.~Ramos, ``Continuous state-action-observation {POMDP}s for trajectory planning with {B}ayesian optimisation,'' in \emph{IEEE/RSJ International Conference on Intelligent Robots and Systems (IROS)}, 2018, pp. 8779--8786.

\bibitem[Fern{\'a}ndez et~al.(2022)Fern{\'a}ndez, Denniston, Caron, and Sukhatme]{fernandez2022informative}
I.~M.~R. Fern{\'a}ndez, C.~E. Denniston, D.~A. Caron, and G.~S. Sukhatme, ``Informative path planning to estimate quantiles for environmental analysis,'' \emph{IEEE Robotics and Automation Letters}, 2022.

\bibitem[R{\"u}ckin et~al.(2022)R{\"u}ckin, Jin, and Popovi{\'c}]{ruckin2022adaptive}
J.~R{\"u}ckin, L.~Jin, and M.~Popovi{\'c}, ``Adaptive informative path planning using deep reinforcement learning for {UAV}-based active sensing,'' in \emph{IEEE International Conference on Robotics and Automation (ICRA)}, 2022.

\bibitem[Ott et~al.(2023)Ott, Balaban, and Kochenderfer]{ott2023sequential}
J.~Ott, E.~Balaban, and M.~J. Kochenderfer, ``Sequential {B}ayesian optimization for adaptive informative path planning with multimodal sensing,'' in \emph{IEEE International Conference on Robotics and Automation (ICRA)}.\hskip 1em plus 0.5em minus 0.4em\relax IEEE, 2023, pp. 7894--7901.

\bibitem[Choudhury et~al.(2020)Choudhury, Gruver, and Kochenderfer]{choudhury2020adaptive}
S.~Choudhury, N.~Gruver, and M.~J. Kochenderfer, ``Adaptive informative path planning with multimodal sensing,'' in \emph{International Conference on Automated Planning and Scheduling (ICAPS)}, vol.~30, 2020, pp. 57--65.

\bibitem[Kochenderfer and Wheeler(2019)]{kochenderfer2019algorithms}
M.~J. Kochenderfer and T.~A. Wheeler, \emph{Algorithms for {O}ptimization}.\hskip 1em plus 0.5em minus 0.4em\relax MIT Press, 2019.

\bibitem[Miller et~al.(1960)Miller, Tucker, and Zemlin]{miller1960integer}
C.~E. Miller, A.~W. Tucker, and R.~A. Zemlin, ``Integer programming formulation of traveling salesman problems,'' \emph{Journal of the ACM (JACM)}, vol.~7, no.~4, pp. 326--329, 1960.

\bibitem[Boyd and Vandenberghe(2004)]{boyd2004convex}
S.~P. Boyd and L.~Vandenberghe, \emph{Convex Optimization}.\hskip 1em plus 0.5em minus 0.4em\relax Cambridge University Press, 2004.

\bibitem[Vansteenwegen et~al.(2011)Vansteenwegen, Souffriau, and Van~Oudheusden]{vansteenwegen2011orienteering}
P.~Vansteenwegen, W.~Souffriau, and D.~Van~Oudheusden, ``The orienteering problem: A survey,'' \emph{European Journal of Operational Research}, vol. 209, no.~1, pp. 1--10, 2011.

\bibitem[Gunawan et~al.(2016)Gunawan, Lau, and Vansteenwegen]{gunawan2016orienteering}
A.~Gunawan, H.~C. Lau, and P.~Vansteenwegen, ``Orienteering problem: A survey of recent variants, solution approaches and applications,'' \emph{European Journal of Operational Research}, vol. 255, no.~2, pp. 315--332, 2016.

\bibitem[Diamond et~al.(2018)Diamond, Takapoui, and Boyd]{diamond2018general}
S.~Diamond, R.~Takapoui, and S.~Boyd, ``A general system for heuristic minimization of convex functions over non-convex sets,'' \emph{Optimization Methods and Software}, vol.~33, no.~1, pp. 165--193, 2018.

\bibitem[Cao et~al.(2022)Cao, Wang, Vashisth, Fan, and Sartoretti]{cao2022catnipp}
Y.~Cao, Y.~Wang, A.~Vashisth, H.~Fan, and G.~Sartoretti, ``Context-aware attention-based network for informative path planning,'' in \emph{6th Annual Conference on Robot Learning}, 2022.

\bibitem[Thomas and Joy(2006)]{thomas2006elements}
M.~Thomas and A.~T. Joy, \emph{Elements of information theory}.\hskip 1em plus 0.5em minus 0.4em\relax Wiley-Interscience, 2006.

\end{thebibliography}

\onecolumn
\appendix
\section{Additional Experiments}
\begin{figure}[t]
\centering
    {\includegraphics[width=1\columnwidth]{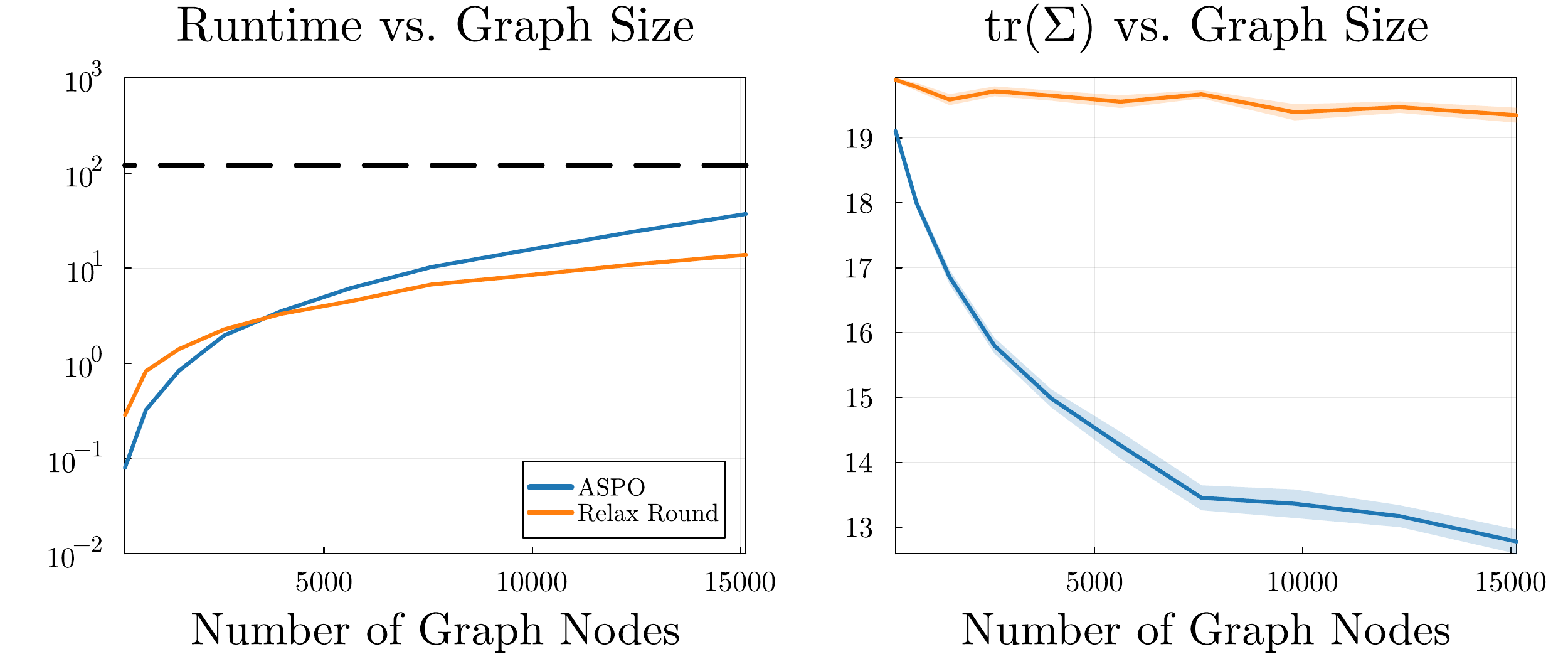}}    
  \caption{Runtime and A-IPP objective as a function of graph size for ASPO and the relax and rounded solution. Each curve shows the average runtime and objective value respectively with the standard error reported over 25 different simulations. The dashed line on the left indicates the 120 second runtime constraint imposed on all methods.} \label{fig:relax_round_vs_aspo_a-ipp}
\end{figure}

\begin{figure}[t]
\centering
    {\includegraphics[width=1\columnwidth]{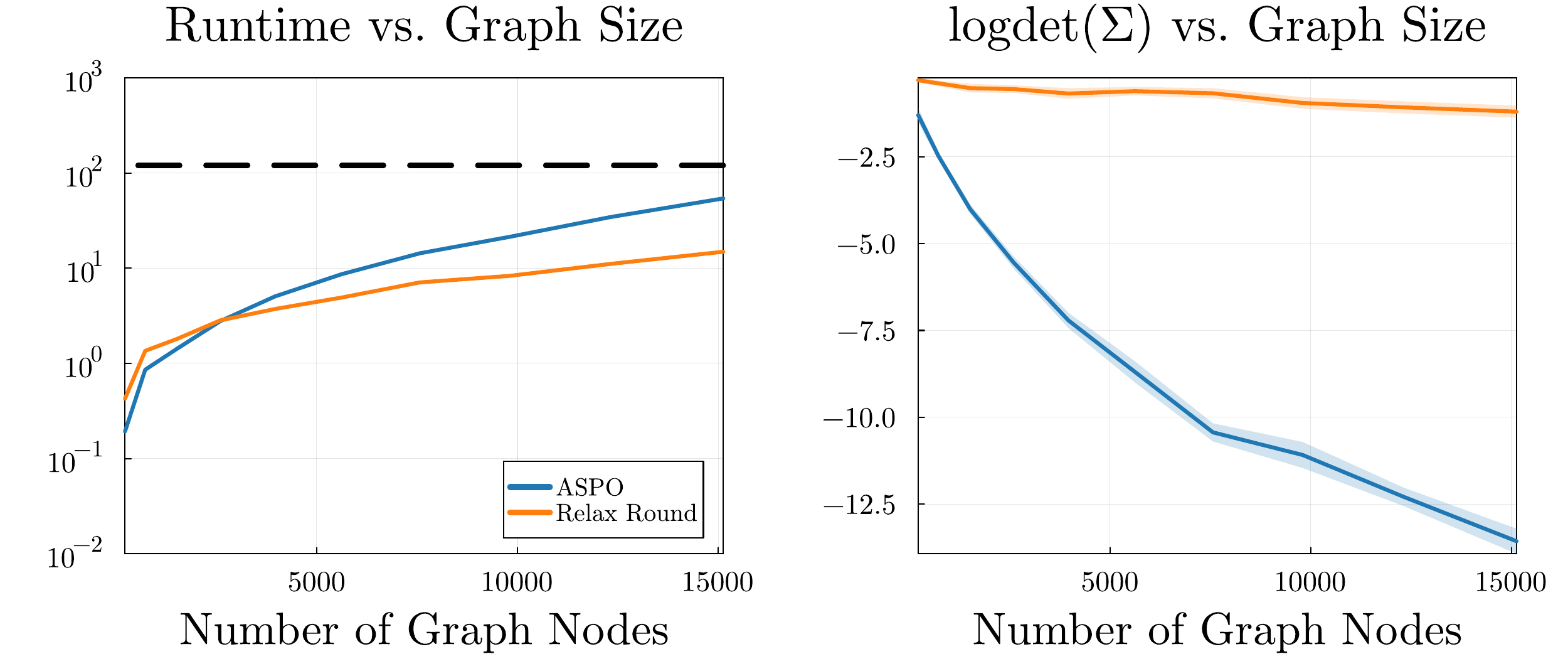}}    
  \caption{Runtime and D-IPP objective as a function of graph size for ASPO and the relax and rounded solution. Each curve shows the average runtime and objective value respectively with the standard error reported over 25 different simulations. The dashed line on the left indicates the 120 second runtime constraint imposed on all methods.} \label{fig:relax_round_vs_aspo_d-ipp}
\end{figure}

\textcolor{black}{We also compared our method against a solution method that first solves the convex relaxation presented in \cref{ss:convex_relaxation}. Notice that the solution to the convex relaxation does not produce a valid path; however, we can construct a valid path by rounding the solution. The rounding procedure starts at the starting node and then selects the neighboring node with the minimum $u$ value that is greater than the current nodes $u$ value given by the constraints in \cref{eq:z_sub1}. The results on the A and D-IPP problems are shown in \cref{fig:relax_round_vs_aspo_a-ipp} and \cref{fig:relax_round_vs_aspo_d-ipp}}. 

\section{Derivation of Maximum a Posteriori Estimate} \label{sec:MAP_derivation}
\begin{proof}
Here we derive the maximum a posteriori estimate of $x$ given in equation \cref{eq:gp_mean}. Given the linear measurement model $y_k = a_{i_k}^T x + \nu_{i_k}$, where $a_i$ is a known vector, $x$ is the parameter vector we want to estimate, and $\nu_i$ is additive Gaussian noise with zero mean and variance $\sigma^2$, the likelihood of observing a particular $y_i$ collected along a path with node indices $( i_1, i_2, \dots, i_{L} )$ given $x$ is: \begin{equation}
p(y_i | x) \propto \exp\left(-\frac{(y_i - a_{i_k}^Tx)^2}{2\sigma^2}\right).
\end{equation}

The likelihood of observing the measurements \( y = [y_1, y_2, \dots, y_L]^T \) given \( x \) is:\begin{equation}
p(y | x) = \prod_{k=1}^L p(y_k | x) = \prod_{k=1}^L \mathcal{N}(y_k | a_{i_k}^T x, \sigma^2)
\end{equation}

Since the noise is Gaussian, this can be written as:
\begin{equation}
p(y | x) = \left( \frac{1}{\sqrt{2\pi\sigma^2}} \right)^L \exp \left( -\frac{1}{2\sigma^2} \sum_{k=1}^L (y_k - a_{i_k}^T x)^2 \right)
\end{equation}

The prior distribution over $x$ is assumed to be Gaussian, \(\mathcal{N}(\bar{x}, \Sigma_x)\): \begin{equation}
p(x) \propto \exp\left(-\frac{1}{2} (x - \bar{x})^T \Sigma_x^{-1} (x - \bar{x})\right).
\end{equation}

From Bayes' rule, the posterior distribution \( p(x | y) \) is:
\begin{equation}
p(x | y) \propto p(y | x) p(x)
\end{equation}

Substituting the expressions for the likelihood and the prior:
\begin{equation}
p(x | y) \propto \exp \left( -\frac{1}{2\sigma^2} \sum_{k=1}^L (y_k - a_{i_k}^T x)^2 \right) \cdot \exp \left( -\frac{1}{2} (x - \bar{x})^T \Sigma_x^{-1} (x - \bar{x}) \right)
\end{equation}

The MAP estimate \( \hat{x} \) maximizes the posterior probability. Equivalently, it minimizes the negative log-posterior probability:
\begin{equation}
\hat{x} = \argmin_x \left[ \frac{1}{2\sigma^2} \sum_{k=1}^L (y_k - a_{i_k}^T x)^2 + \frac{1}{2} (x - \bar{x})^T \Sigma_x^{-1} (x - \bar{x}) \right]
\end{equation}

The solution can be found by setting the gradient with respect to \( x \) to zero:
\begin{equation}
\frac{\partial}{\partial x} \left[ \frac{1}{2\sigma^2} \sum_{k=1}^L (y_k - a_{i_k}^T x)^2 + \frac{1}{2} (x - \bar{x})^T \Sigma_x^{-1} (x - \bar{x}) \right] = 0
\end{equation}

\begin{equation}
-\frac{1}{\sigma^2} \sum_{k=1}^L a_{i_k} (y_k - a_{i_k}^T x) + \Sigma_x^{-1} (x - \bar{x}) = 0
\end{equation}

Simplifying:

\begin{equation}
\frac{1}{\sigma^2} \sum_{k=1}^L a_{i_k} a_{i_k}^T x - \frac{1}{\sigma^2} \sum_{k=1}^L a_{i_k} y_k + \Sigma_x^{-1} x - \Sigma_x^{-1} \bar{x} = 0
\end{equation}

\begin{equation}
\left( \frac{1}{\sigma^2} \sum_{k=1}^L a_{i_k} a_{i_k}^T + \Sigma_x^{-1} \right) x = \frac{1}{\sigma^2} \sum_{k=1}^L a_{i_k} y_k + \Sigma_x^{-1} \bar{x}
\end{equation}

Rearranging:
\begin{equation}
\hat{x} = \left( \sigma^{-2} \sum_{k=1}^L a_{i_k} a_{i_k}^T + \Sigma_x^{-1} \right)^{-1} \left( \sigma^{-2} \sum_{k=1}^L a_{i_k} y_k + \Sigma_x^{-1} \bar{x} \right)
\end{equation}

Distributing the first term:
\begin{equation}
\hat{x} = \left( \sigma^{-2} \sum_{k=1}^L a_{i_k} a_{i_k}^T + \Sigma_x^{-1} \right)^{-1} \left( \sigma^{-2} \sum_{k=1}^L a_{i_k} y_k \right) + 
\left( \sigma^{-2} \sum_{k=1}^L a_{i_k} a_{i_k}^T + \Sigma_x^{-1} \right)^{-1} \Sigma_x^{-1} \bar{x} \label{eq:appendix_A_chkpt1}
\end{equation}

We can rewrite the second term as follows: 
\begin{equation}
\begin{aligned}
   &\left( \sigma^{-2} \sum_{k=1}^L a_{i_k} a_{i_k}^T + \Sigma_x^{-1} \right)^{-1} \Sigma_x^{-1} \bar{x} \\
   &= \left( \sigma^{-2} \sum_{k=1}^L a_{i_k} a_{i_k}^T + \Sigma_x^{-1} \right)^{-1} \left( \sigma^{-2} \sum_{k=1}^L a_{i_k} a_{i_k}^T \bar{x} - \sigma^{-2} \sum_{k=1}^L a_{i_k} a_{i_k}^T \bar{x} + \Sigma_x^{-1} \bar{x} \right) \\
   &= \left( \sigma^{-2} \sum_{k=1}^L a_{i_k} a_{i_k}^T + \Sigma_x^{-1} \right)^{-1} \left( - \sigma^{-2} \sum_{k=1}^L a_{i_k} a_{i_k}^T \bar{x} + \left( \sigma^{-2} \sum_{k=1}^L a_{i_k} a_{i_k}^T + \Sigma_x^{-1} \right) \bar{x} \right) \\
   &= \left( \sigma^{-2} \sum_{k=1}^L a_{i_k} a_{i_k}^T + \Sigma_x^{-1} \right)^{-1} \left( - \sigma^{-2} \sum_{k=1}^L a_{i_k} a_{i_k}^T \bar{x} \right) + \bar{x}
\end{aligned}
\end{equation}

Substituting this back into \cref{eq:appendix_A_chkpt1}:
\begin{equation}
\begin{aligned}
\hat{x} &= \left( \sigma^{-2} \sum_{k=1}^L a_{i_k} a_{i_k}^T + \Sigma_x^{-1} \right)^{-1} \left( \sigma^{-2} \sum_{k=1}^L a_{i_k} y_k \right) \\
 &+ \left( \sigma^{-2} \sum_{k=1}^L a_{i_k} a_{i_k}^T + \Sigma_x^{-1} \right)^{-1} \left( - \sigma^{-2}\sum_{k=1}^L a_{i_k} a_{i_k}^T \bar{x} \right) + \bar{x} \\
&= \left( \sigma^{-2} \sum_{k=1}^L a_{i_k} a_{i_k}^T + \Sigma_x^{-1} \right)^{-1} \left( \sigma^{-2} \sum_{k=1}^L \left( a_{i_k} y_k - a_{i_k} a_{i_k}^T \bar{x} \right) \right) + \bar{x}
\end{aligned}
\end{equation}

Recall $\bar{y}_k = a_{i_k}^T \bar{x}$:

\begin{equation}
\hat{x} = \left( \sigma^{-2} \sum_{k=1}^L a_{i_k} a_{i_k}^T + \Sigma_x^{-1} \right)^{-1} \left( \sigma^{-2} \sum_{k=1}^L a_{i_k} y_k - a_{i_k} \bar{y}_k \right) + \bar{x}
\end{equation}

Finally: 
\begin{equation}
\hat{x} = \bar{x} + \left( \sigma^{-2} \sum_{k=1}^L a_{i_k} a_{i_k}^T + \Sigma_x^{-1} \right)^{-1} \sigma^{-2} \sum_{k=1}^L (y_k - \bar{y}_k)a_{i_k}
\end{equation} 
\end{proof}

\section{Derivation of Posterior Covariance} \label{sec:MAP_covariance_derivation}
\begin{proof}
The covariance of the posterior distribution is the inverse of the Hessian of the negative log-posterior with respect to $x$, evaluated at the MAP estimate. The Hessian can be found starting from the expression for the negative log-posterior

\begin{equation}
-\log p(x | y) = \frac{1}{2\sigma^2} \sum_{k=1}^L (y_k - a_{i_k}^T x)^2 + \frac{1}{2} (x - \bar{x})^T \Sigma_x^{-1} (x - \bar{x}) + \text{const}
\end{equation}

Computing the Hessian:
\begin{equation}
H = \frac{\partial^2}{\partial x^2} \left( \frac{1}{2\sigma^2} \sum_{k=1}^L (y_k - a_{i_k}^T x)^2 + \frac{1}{2} (x - \bar{x})^T \Sigma_x^{-1} (x - \bar{x}) \right)
\end{equation}

The second derivative of the first term is:
\begin{equation}
\frac{\partial^2}{\partial x^2} \left( \frac{1}{2\sigma^2} \sum_{k=1}^L (y_k - a_{i_k}^T x)^2 \right) = \frac{1}{\sigma^2} \sum_{k=1}^L a_{i_k} a_{i_k}^T
\end{equation}

The second derivative of the second term is:
\begin{equation}
\frac{\partial^2}{\partial x^2} \left( \frac{1}{2} (x - \bar{x})^T \Sigma_x^{-1} (x - \bar{x}) \right) = \Sigma_x^{-1}
\end{equation}

Thus, the Hessian is:
\begin{equation}
H = \frac{1}{\sigma^2} \sum_{k=1}^L a_{i_k} a_{i_k}^T + \Sigma_x^{-1}
\end{equation}

The covariance matrix of the posterior distribution is the inverse of the Hessian:
\begin{equation}
\Sigma = \left( \sigma^{-2} \sum_{k=1}^L a_{i_k} a_{i_k}^T + \Sigma_x^{-1} \right)^{-1}
\end{equation}

which is \cref{eq:gp_var}. 
\end{proof}

\section{Gaussian Process Equivalence} \label{sec:gp_equivalence}
A Gaussian process is a distribution over functions that can be used to predict an underlying function $f$ given some previously observed noisy measurements at location $\textbf{x}$. That is, $y = f(\textbf{x}) + z$ where $f$ is deterministic but $z$ is zero-mean Gaussian noise, $z \sim \mathcal{N}(0, \sigma^2)$. The distribution of functions conditioned on the previously observed measurements  $\boldsymbol{\hat{y}} \mid \boldsymbol{y}, \sigma^2 \sim \mathcal{N}(\boldsymbol{\mu^{\ast}}, \boldsymbol{\Sigma^{\ast}})$ is given by: \begin{gather}
    \boldsymbol{\mu^{\ast}} = \boldsymbol{m}(X^{\ast})
    \newline + \boldsymbol{K}(X^{\ast}, X)(\boldsymbol{K}(X, X) + \sigma^2 \boldsymbol{I})^{-1} (\boldsymbol{y} - \boldsymbol{m}(X)) \label{eq:appendix_gp_mean} \\
    \boldsymbol{\Sigma^{\ast}} = \boldsymbol{K}(X^{\ast}, X^{\ast}) - \boldsymbol{K}(X^{\ast}, X)(\boldsymbol{K}(X,X) + \sigma^2 \boldsymbol{I})^{-1} \boldsymbol{K}(X, X^{\ast}) \label{eq:appendix_gp_covar}
\end{gather} where $X$ is the set of previously measured locations, $X^{\ast}$ is the set of locations we wish to predict the values $\boldsymbol{\hat{y}}$, $\boldsymbol{m}(X)$ is the mean function and $\boldsymbol{K}(X,X')$ is the covariance matrix constructed with the kernel $k(\boldsymbol{x},\boldsymbol{x'})$ \cite{kochenderfer2019algorithms}.

We want to show that \cref{eq:appendix_gp_mean} and \cref{eq:appendix_gp_covar} are equivalent to \begin{equation}
    \hat{x} = \bar{x} + \left( \sigma^{-2} \sum_{k=1}^L a_{i_k} a_{i_k}^T + \Sigma_x^{-1} \right)^{-1} \sigma^{-2} \sum_{k=1}^L (y_k - \bar{y}_k)a_{i_k},
\end{equation} and \begin{equation}
\Sigma = \left( \sigma^{-2}  \sum_{k=1}^L a_{i_k} a_{i_k}^T + \Sigma_x^{-1} \right)^{-1}
\end{equation} from \cref{eq:gp_mean} and \cref{eq:gp_var} respectively. 

\subsection{Covariance Equivalence}

\begin{proof}
We know that each $a_i \in \mathbf{R}^{m}$ characterizes the measurements and that there are $n$ unique $a_i$ corresponding to each of the $n$ locations in the environment. In the Gaussian process formulation, $X$ denotes the previously measured locations and therefore corresponds to the path indices $p = ( i_1, i_2, \dots, i_{L} )$. Then \begin{equation}
A = \begin{bmatrix}
    - a_{i_1}^T - \\ 
    - a_{i_2}^T - \\
    \vdots \\
    - a_{i_L}^T -
\end{bmatrix}
\end{equation} with $A \in \mathbf{R}^{L \times m}$. Notice that $\sum_{k=1}^L a_{i_k} a_{i_k}^T = A^T A$ with $A^T A \in \mathbf{R}^{m \times m}$.

For equivalence, we have that the query points $X^{\ast}$ correspond to the locations in the environment that we want to estimate the vector $x$ at.\footnote{Notice that $X^{\ast}$ correspond to locations in the environment while $x$ represents the vector we wish to estimate, i.e. the values at those locations in the environment.} Then we have the following equalities: 
\begin{enumerate}
    \item  $m(X^{\ast}) = \bar{x}$
    \item $\boldsymbol{K}(X^{\ast}, X^{\ast}) = \Sigma_x$
    \item $\boldsymbol{K}(X^{\ast}, X) = \Sigma_x A^T$
\end{enumerate} where the last equality is specified by the sensor characterization. 

Using the Woodbury matrix identity\footnote{The Woodbury matrix identity: $\left(A + UCV \right)^{-1} = A^{-1} - A^{-1}U \left(C^{-1} + VA^{-1}U \right)^{-1} VA^{-1}$} we can rewrite \cref{eq:appendix_gp_covar} as \begin{equation}
    \boldsymbol{\Sigma^{\ast}} = \left( \boldsymbol{K}(X^{\ast}, X^{\ast})^{-1} + \boldsymbol{K}(X^{\ast}, X^{\ast})^{-1} \boldsymbol{K}(X^{\ast}, X) \sigma^{-2} \boldsymbol{I} \boldsymbol{K}(X, X^{\ast}) \boldsymbol{K}(X^{\ast}, X^{\ast})^{-1} \right)^{-1}
\end{equation}

Substituting in equalities $2$ and $3$: \begin{equation}
\begin{aligned}
    \boldsymbol{\Sigma^{\ast}} &= \left( \Sigma_x^{-1} + \sigma^{-2} \Sigma_x^{-1} \boldsymbol{K}(X^{\ast}, X) \boldsymbol{K}(X, X^{\ast}) \Sigma_x^{-1} \right)^{-1} \\
    &= \left( \Sigma_x^{-1} + A^T A \right)^{-1}  \\
    &= \left( \sigma^{-2} \sum_{k=1}^L a_{i_k} a_{i_k}^T + \Sigma_x^{-1} \right)^{-1} \\
    &= \Sigma 
\end{aligned}
\end{equation}
\end{proof}

\subsection{Mean Equivalence}
\begin{proof}
Starting from the expression for the Gaussian process mean and using the previous result from Woodbury matrix identity\footnote{Notice we had $\boldsymbol{K}(X, X) = VA^{-1}U = \boldsymbol{K}(X, X^{\ast})\boldsymbol{K}(X^{\ast}, X^{\ast})^{-1} \boldsymbol{K}(X^{\ast}, X)$ in the Woodbury matrix identity.}: \begin{equation}
\begin{aligned}
    \boldsymbol{\mu^{\ast}} &= \boldsymbol{m}(X^{\ast}) + \boldsymbol{K}(X^{\ast}, X)(\boldsymbol{K}(X, X) + \sigma^2 \boldsymbol{I})^{-1} (\boldsymbol{y} - \boldsymbol{m}(X)) \\
    &= \boldsymbol{m}(X^{\ast})
    \newline + \boldsymbol{K}(X^{\ast}, X)(\boldsymbol{K}(X, X^{\ast})\boldsymbol{K}(X^{\ast}, X^{\ast})^{-1} \boldsymbol{K}(X^{\ast}, X) + \sigma^2 \boldsymbol{I})^{-1} (\boldsymbol{y} - \boldsymbol{m}(X))
\end{aligned}
\end{equation} 

Substituting in $\boldsymbol{K}(X^{\ast}, X) = \Sigma_x A^T$ and $\boldsymbol{K}(X, X^{\ast}) = A \Sigma_x $: \begin{equation}
\begin{aligned}
    \boldsymbol{\mu^{\ast}} &= \boldsymbol{m}(X^{\ast}) + \Sigma_x A^T (A \Sigma_x \Sigma_x^{-1} \Sigma_x A^T + \sigma^2 \boldsymbol{I})^{-1} (\boldsymbol{y} - \boldsymbol{m}(X)) \\
    &= \boldsymbol{m}(X^{\ast}) + \Sigma_x A^T (A \Sigma_x A^T + \sigma^2 \boldsymbol{I})^{-1} (\boldsymbol{y} - \boldsymbol{m}(X)) \label{eq:appendix_mean_equiv_chkpt1}
\end{aligned}
\end{equation} 

Notice that $\Sigma_x A^T (A \Sigma_x A^T + \sigma^2 \boldsymbol{I})^{-1} = (A^T\sigma^{-2} \boldsymbol{I}A +\Sigma_x^{-1})^{-1} A^T\sigma^{-2} \boldsymbol{I}$. To see this, multiply \begin{equation}
    \Sigma_x A^T (A \Sigma_x A^T + \sigma^2 \boldsymbol{I})^{-1} \stackrel{?}{=} (A^T\sigma^{-2} \boldsymbol{I}A + \Sigma_x^{-1})^{-1} A^T\sigma^{-2} \boldsymbol{I}
\end{equation} 

on the left by $(A^T\sigma^{-2} \boldsymbol{I}A + \Sigma_x^{-1})$ and on the right by $(A \Sigma_x A^T + \sigma^2 \boldsymbol{I})$ to get \begin{equation}
     (A^T\sigma^{-2} \boldsymbol{I}A + \Sigma_x^{-1}) \Sigma_x A^T \stackrel{?}{=}A^T\sigma^{-2} \boldsymbol{I}(A \Sigma_x A^T + \sigma^2 \boldsymbol{I}) \\
\end{equation} 

which is true.

Now substituting in $(A^T\sigma^{-2} \boldsymbol{I}A +\Sigma_x^{-1})^{-1} A^T\sigma^{-2} \boldsymbol{I}$ for $\Sigma_x A^T (A \Sigma_x A^T + \sigma^2 \boldsymbol{I})^{-1}$ in \cref{eq:appendix_mean_equiv_chkpt1} we get: \begin{equation}
\begin{aligned}
    \boldsymbol{\mu^{\ast}} &= \boldsymbol{m}(X^{\ast}) +  (A^T\sigma^{-2} \boldsymbol{I}A +\Sigma_x^{-1})^{-1} A^T\sigma^{-2} \boldsymbol{I} (\boldsymbol{y} - \boldsymbol{m}(X)) \\
    &= \bar{x} +  (\sigma^{-2}A^TA +\Sigma_x^{-1})^{-1} \sigma^{-2} A^T (\boldsymbol{y} - \boldsymbol{m}(X)) \\
    &= \bar{x} + (\sigma^{-2} \sum_{k=1}^L a_{i_k} a_{i_k}^T + \Sigma_x^{-1})^{-1} \sigma^{-2} \sum_{k=1}^L (y_k - \bar{y}_k)a_{i_k} \\
    &= \hat{x}
\end{aligned}
\end{equation}
\end{proof}

\section{Mutual Information} \label{sec:mutual_information}
\textcolor{black}{Starting from the Gaussian process formulation provided by \cref{eq:appendix_gp_mean} and \cref{eq:appendix_gp_covar} we have that the distribution of functions conditioned on the previously observed measurements  $\boldsymbol{\hat{y}} \mid \boldsymbol{y}, \sigma^2 \sim \mathcal{N}(\boldsymbol{\mu^{\ast}}, \boldsymbol{\Sigma^{\ast}})$ is given by: \begin{gather}
    \boldsymbol{\mu^{\ast}} = \boldsymbol{m}(X^{\ast})
    \newline + \boldsymbol{K}(X^{\ast}, X)(\boldsymbol{K}(X, X) + \sigma^2 \boldsymbol{I})^{-1} (\boldsymbol{y} - \boldsymbol{m}(X))  \\
    \boldsymbol{\Sigma^{\ast}} = \boldsymbol{K}(X^{\ast}, X^{\ast}) - \boldsymbol{K}(X^{\ast}, X)(\boldsymbol{K}(X,X) + \sigma^2 \boldsymbol{I})^{-1} \boldsymbol{K}(X, X^{\ast}). 
\end{gather} For a multivariate normal distribution \( \mathbf{z} \sim \mathcal{N}(0, \mathbf{P}) \) where $z \in \mathbf{R}^n$, the differential entropy is given by: \begin{equation}
    H(\mathbf{z}) = \frac{1}{2} \log \left( (2\pi e)^n \det(\mathbf{P}) \right).
\end{equation} For the conditional distribution \( \hat{\mathbf{y}} \mid \mathbf{y} \), which is Gaussian with covariance \( \boldsymbol{\Sigma}^* \), the conditional entropy is: \begin{equation}
H(\hat{\mathbf{y}} \mid \mathbf{y}) = \frac{1}{2} \log \left( (2\pi e)^{m} \det(\boldsymbol{\Sigma}^*) \right)
\end{equation} since we are interested in making predictions at the $m$ locations in the environment \cite{thomas2006elements}.}

\textcolor{black}{The mutual information between the observed outputs \( \mathbf{y} \) and the predicted outputs \( \hat{\mathbf{y}} \) is given by: \begin{equation}
    I(\hat{\mathbf{y}}; \mathbf{y}) = H(\hat{\mathbf{y}}) - H(\hat{\mathbf{y}} \mid \mathbf{y}). 
\end{equation} Given that \( \hat{\mathbf{y}} \) and \( \mathbf{y} \) have a joint Gaussian distribution, the marginal entropy \( H(\hat{\mathbf{y}}) \) can be computed similarly: \begin{equation}
    H(\hat{\mathbf{y}}) = \frac{1}{2} \log \left( (2\pi e)^m \det(\mathbf{K}(X^*, X^*)) \right)
\end{equation} Therefore, the mutual information is: \begin{equation}
    I(\hat{\mathbf{y}}; \mathbf{y}) = \frac{1}{2} \log \left( \frac{\det(\mathbf{K}(X^*, X^*))}{\det(\boldsymbol{\Sigma}^*)} \right) = \frac{1}{2} \log \left(\det(\mathbf{K}(X^*, X^*)) \right) - \frac{1}{2} \log \left( \det(\boldsymbol{\Sigma}^*) \right) . 
\end{equation} Rewriting this using the notation from \cref{sec:ipp} we have: \begin{equation}
    I(\hat{x}; \{y_1, \dots, y_L \}) = \frac{1}{2} \log \left( \frac{\det(\Sigma_x)}{\det(\Sigma)} \right) = \frac{1}{2} \log \left(\det(\Sigma_x) \right) - \frac{1}{2} \log \left( \det(\Sigma) \right). 
\end{equation} Therefore maximizing the mutual information is equivalent to minimizing $\log \left( \det(\Sigma) \right)$. }

\label{sec:appendix_A}

\end{document}